%% file: preprint.tex
\PassOptionsToPackage{numbers,sort&compress,square,comma}{natbib}
\documentclass{article}

\usepackage[eandd, preprint]{neurips_2026}

\usepackage[utf8]{inputenc}
\usepackage[T1]{fontenc}
\usepackage{hyperref}
\usepackage{url}
\usepackage{booktabs}
\usepackage{amsfonts}
\usepackage{nicefrac}
\usepackage{microtype}
\usepackage{xcolor}
\usepackage{graphicx}
\usepackage{tabularx}
\usepackage{array}
\usepackage{makecell}
\usepackage{enumitem}
\usepackage{multirow}
\usepackage{subcaption}
\usepackage{longtable}
\setcitestyle{numbers,square,comma,sort&compress}
\usepackage{float}

\title{Beyond Accuracy: Evaluating Strategy Diversity in LLM Mathematical Reasoning}

\author{
Xia Yang$^{1}$ \qquad
Xuanyi Zhang$^{2}$ \qquad
Hao Hu$^{3}$ \qquad
Feng Ji$^{1}$\thanks{Corresponding author: \texttt{f.ji@utoronto.ca}} \\
$^{1}$University of Toronto,\qquad
$^{2}$Upper Canada College,\qquad
$^{3}$East China Normal University
}

\newcommand{\fref}[1]{Figure~\ref{#1}}
\newcommand{\tref}[1]{Table~\ref{#1}}

\begin{document}

\maketitle

\begin{abstract}
Large language models now achieve high final-answer accuracy on mathematical reasoning benchmarks, but accuracy alone does not capture reasoning flexibility. We introduce a strategy-level evaluation framework instantiated on 80 AMC 10/12 and AIME problems with 217 AoPS-derived reference strategy families. Model outputs are annotated for strategy identity, validity, and correctness using dual-AI coding with human adjudication. Across four frontier models, we find a pronounced decoupling between answer accuracy and strategy diversity. Under a single-solution prompt, all models achieve high accuracy (95\%–100\%), but under a multiple-strategy prompt they recover substantially fewer strategies than the human reference set. Gemini, DeepSeek, GPT, and Claude generate 184, 152, 151, and 110 distinct valid strategies, respectively, with the largest gaps in Geometry and Number Theory. The models collectively produce 50 benchmark-novel valid strategies, indicating both incomplete coverage of human strategies and some capacity for alternative reasoning. A repeated-run robustness check on 20 problems shows diminishing gains in discovered strategies, with the strongest model recovering only 39 of 55 AoPS-reference strategies (71\%) after three runs. These findings position strategy diversity as a complementary dimension for evaluating mathematical reasoning beyond answer correctness.
\end{abstract}

\section{Introduction}
Mathematical reasoning has long been regarded as a central benchmark for artificial general intelligence, reflecting a unique synthesis of formal logic, symbolic manipulation, and creative insight \citep{polya1945how,newell1956logic}. In recent years, large language models (LLMs) have advanced rapidly in this domain, with progress driven by both math-specialized language models and neuro-symbolic systems for mathematical problem solving \citep{deepseek2025v32,trinh2024solving}. Driven by specialized continued pre-training on large-scale mathematical corpora \citep{azerbayev2024llemma}, process-level supervision \citep{lightman2023lets}, and hybrid reasoning methods that combine language models with symbolic engines or tree-based search \citep{trinh2024solving,yao2023tree}, recent systems have made rapid progress on mathematical reasoning tasks. The benchmark landscape now spans arithmetic and algebraic word problems \citep{cobbe2021gsm8k,hendrycks2021measuring} , competition mathematics \citep{ling2017program, gao2024omnimath,he2024olympiadbench,zheng2021minif2f,sun2025challenging}, and advanced mathematical reasoning tasks such as research-level problem solving and formal theorem proving  \citep{garre2026riemannbench, azerbayev2023proofnet,chaudhuri2024putnambench,liu2023fimo,zhang2025realmath}. Yet, even as evaluation methods move beyond final-answer accuracy to consider intermediate reasoning steps and metacognitive traces \citep{lightman2023lets,arora2024metacognitive}, and as prompting and search methods such as chain-of-thought prompting, self-consistency sampling, and tree-based reasoning enable models to generate explicit or multiple reasoning trajectories \citep{yao2023tree,wang2022selfconsistency,wei2022cot}, evaluation remains largely centered on validating individual trajectories rather than assessing the diversity of alternative solution strategies. This focus on one-path success raises a fundamental question: does a model's success on complex problems reflect robust mathematical understanding, or merely optimized search within a narrow reasoning corridor?

In human practice, mathematical proficiency is characterized by strategic flexibility: the ability to perceive a mathematical object through multiple lenses \citep{leikin2009creativity,leikin2013stateofart,rittlejohnson2020multiple} and transition fluently among divergent approaches and semiotic registers \citep{duval2006cognitive,duval2017registers}. Such flexibility depends not only on the availability of resources (i.e., domain knowledge), but also on a high level of metacognitive control to coordinate heuristics and navigate competing solution paths, while attending to the underlying structural relationships that organize the problem space \citep{schoenfeld2016learning}. As Klowden and Tao emphasize, in the age of AI it is essential to consider how computational methods relate to human mathematical thought and practice \citep{mathematicalmethods2026}. The mismatch between these accounts of human mathematical practice and current AI evaluation frameworks exposes a critical gap: existing assessments emphasize correctness but offer limited insight into the strategic diversity and structural depth of reasoning.

This limitation has implications beyond model evaluation, particularly for the design of AI systems that interact with human problem-solving processes. The effectiveness of such systems depends on their ability to provide step-level guidance that is responsive to a learner's evolving solution path \citep{kulik2016its}. To function as genuinely adaptive partners, large language models must therefore exhibit sufficient strategic breadth to recognize and navigate the diverse solution topologies that arise in human reasoning \citep{lake2016people,stamper2024feedback}.

\begin{figure*}[t]
  \centering
  \includegraphics[width=0.80\textwidth]{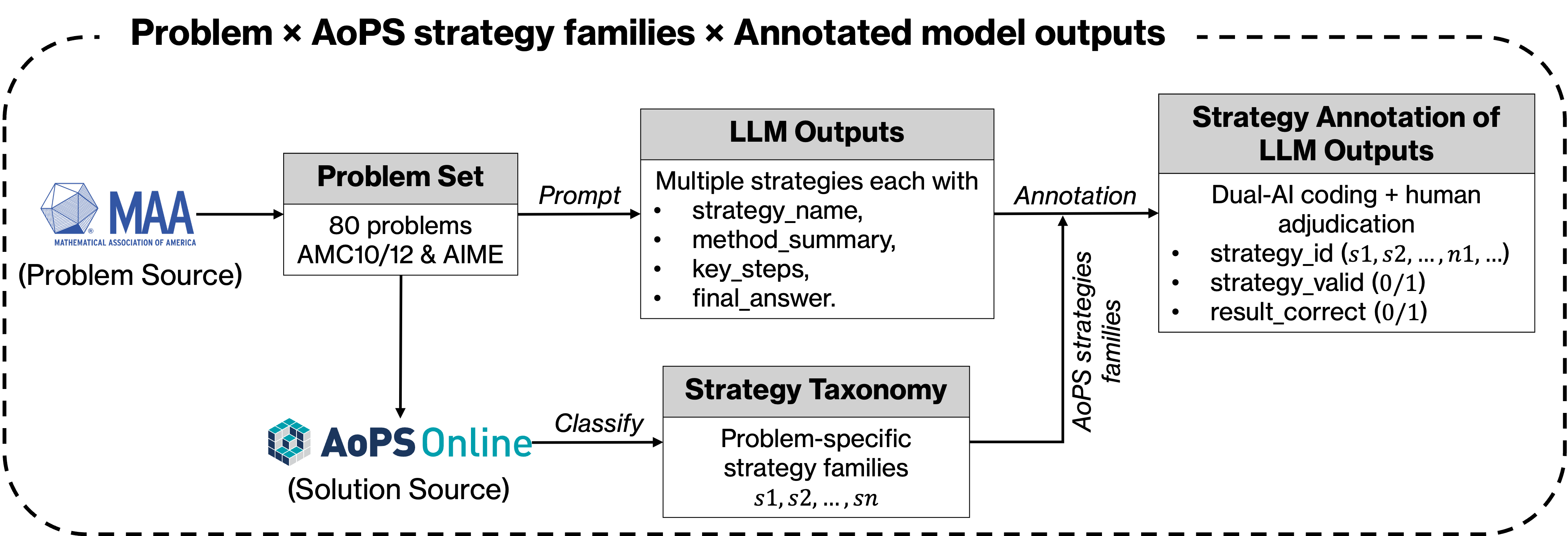}
  \caption{Overview of the dataset construction and evaluation pipeline.}
  \label{fig:workflow}
\end{figure*}

This motivates a broader paradigm in which reasoning systems are evaluated by the richness and organization of the reasoning processes they exhibit. We propose a strategy-level evaluation framework that treats strategy diversity as a key dimension of mathematical reasoning. The overall workflow is shown in \fref{fig:workflow}. We instantiate this framework as a strategy-annotated evaluation dataset built over 80 competition problems drawn from AMC10, AMC12 and AIME \citep{maa}. The 80 problems covers a total of 217 reference strategy families derived from the Art of Problem Solving (AoPS) \citep{aops}. The unit of analysis is therefore problem-specific strategy space, together with model-output annotations for strategy family, validity, and correctness. Because constructing such strategy taxonomies requires substantial expert mathematical coding, this instantiation emphasizes annotation depth rather than large-scale problem coverage. 

We evaluate four frontier LLMs not primarily to produce a leaderboard, but to characterize how current reasoning models behave under this strategy-level evaluation lens. We ask: when a model solves a problem multiple times, does it explore the breadth of the mathematical solution space, or does it remain confined to a narrow, repetitive corridor of reasoning? Using a dual-AI coding pipeline with targeted human adjudication, we move beyond the “Pass@k” paradigm and show that while modern systems achieve high final-answer accuracy, they recover substantially fewer distinct reasoning strategies than their human counterparts, with particularly large gaps in Geometry and Number Theory. While models occasionally discover valid novel approaches not present in the reference corpus, repeated sampling yields diminishing returns in strategic variety. Our contributions are as follows:

\textbf{Formalizing strategic diversity.} We introduce strategy diversity as a core dimension for evaluating mathematical reasoning systems. By shifting focus from a single-solution path to the breadth of a model's reasoning repertoire, we provide a principled lens for assessing generalized mathematical competence.

\textbf{An augmented evaluation framework.} We develop a systematic methodology for constructing a strategy-annotated evaluation dataset, combining a problem-specific taxonomy derived from expert human solvers (AoPS) with a scalable dual-AI verification pipeline.

\textbf{The diversity-accuracy decoupling.} We provide, to our knowledge, the first systematic evidence that while frontier models approach human-level final-answer accuracy, they recover substantially fewer reasoning strategies than the human reference set. This diversity gap underscores the need for diversity-at-discovery---the capacity to explore varied reasoning topologies.

\section{Related work}
The landscape of AI evaluation has widely centered on Olympiad-level datasets \citep{gao2024omnimath,he2024olympiadbench,zheng2021minif2f,sun2025challenging}, which challenge frontier models with high-difficulty problems but remain fundamentally outcome-oriented, prioritizing final-answer accuracy. Our work aligns with a growing line of research that seeks to move beyond accuracy by examining the structural properties of reasoning, such as reasoning path divergence and problem-aware strategy routing \citep{ju2025reasoningpath,qi2025plan}. We extend this direction by grounding evaluation in a human-expert taxonomy of strategies derived from the AoPS community, thereby linking model behavior to established patterns of expert mathematical practice.

Our operationalization of strategic distinction draws on Roza Leikin's framework for mathematical creativity and flexibility \citep{leikin2009creativity,leikin2013stateofart,levavwaynberg2012role}. We define a distinct strategy not by surface-level variations, but as a solution family governed by a decisive reasoning move that determines the structural form of the argument. In contrast, forms of procedural flexibility, such as those observed in equation-solving contexts \citep{star2008flexibility}, do not reach the level of the conceptual strategic variations targeted in this work. By focusing on these structural milestones, our metric captures genuine cognitive shifts rather than superficial differences in algebraic manipulation, computational order, or explanatory style.

\section{Dataset and annotation}
\subsection{Problem selection and strategy sources}
We construct the dataset component of the evaluation framework as a strategy-annotated evaluation set, in which each item comprises a competition-style problem together with a problem-specific reference space of solution strategies. The problem set contains 80 AMC10, AMC12 and AIME problems administered by the MAA. These problems are well-defined, formally verifiable, and frequently admit multiple qualitatively distinct solution approaches, making them suitable for evaluating both answer correctness and strategy diversity. We assign each problem to one of five high-level domains: Algebra (14), Combinatorics (19), Geometry (16), Number Theory (15), and Probability (16). The Geometry subset excludes problems whose statements depend on an accompanying diagram, but includes text-only problems whose solutions still require internal geometric visualization. This distinction avoid confounds introduced by diagram interpretation, which has been identified as a distinct challenge in multimodal mathematical reasoning (e.g., MATH-Vision \citep{wang2024mathvision}). 

For each problem, we collect corresponding human-written solutions from the AoPS community. To avoid biasing the framework toward problems with unusually rich documented solution spaces, we balance problems by the number of solutions available on AoPS. In addition to problems with many known solution methods, we include problems with only a single recorded AoPS solution. This lets us evaluate both whether models can recover known strategies and whether they can generate alternative valid approaches when the human reference is sparse. The released framework dataset, coding files, and metadata are hosted on \href{https://doi.org/10.34740/kaggle/ds/10271409}{Kaggle}.

\subsection{Strategies classification and annotation}

\begin{figure}[tbp]
  \centering
  \includegraphics[width=0.92\columnwidth]{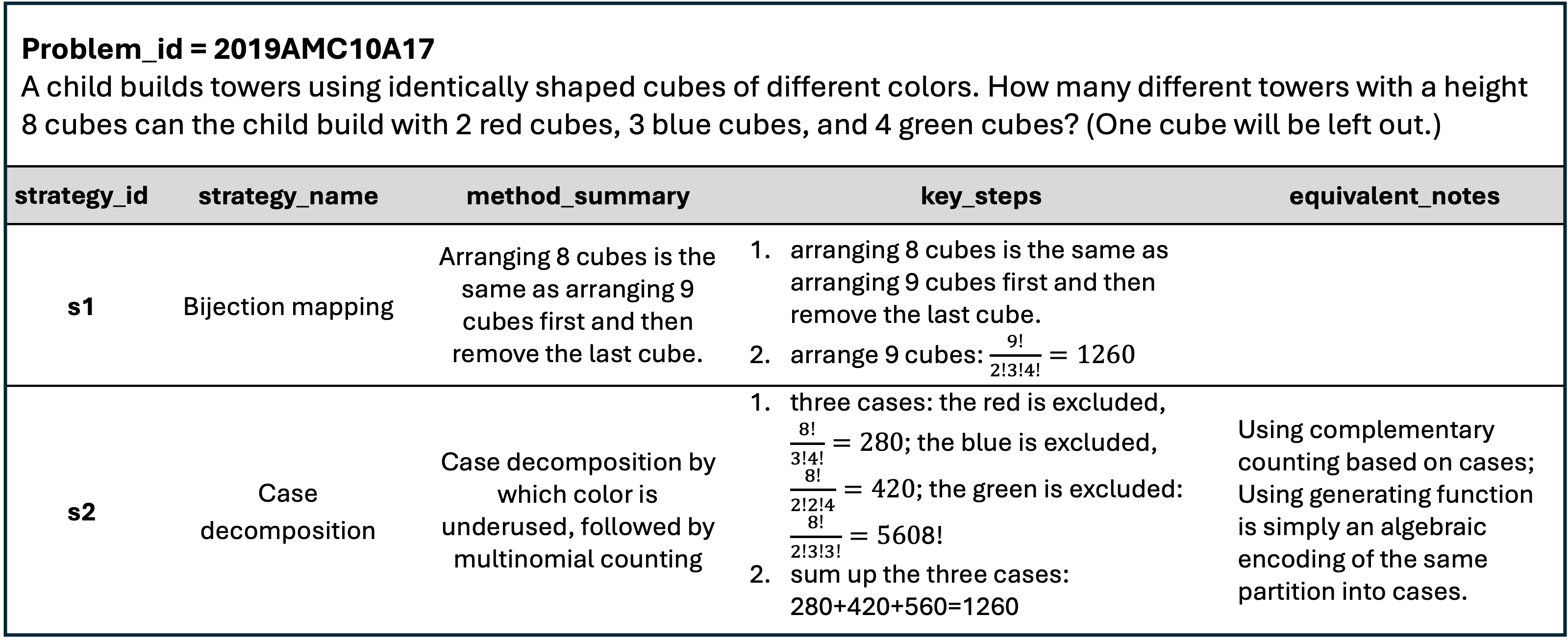}
  \caption{Illustrative example of AoPS-derived reference strategies for a single problem.}
  \label{fig:aopsexample}
\end{figure}

In our classification, the decisive move of problem-solving may manifest as an initial entry-point insight or as a pivotal late-stage step that completes the logical structure of the solution. Strategy labels are defined locally at the problem level rather than globally across the dataset. We do not assume that labels such as casework or coordinate geometry carry uniform meaning across different problems. Instead, for each problem, solutions are grouped according to the substantive reasoning distinctions relevant in that context. For each problem, human solutions from AoPS are grouped into reference strategy families, denoted by \texttt{strategy\_id} values such as $s_1, s_2, \ldots, s_n$.

To establish the reference taxonomy, two independent coder pairs annotated an overlapping subset of 30 problems (14 coded by Coder A/B and 16 by Coder C/D). Because coders did not always use the same provisional strategy names, inter-rater reliability was assessed using temporary within-problem comparison slots that aligned corresponding candidate strategies across coders. On this basis, pooled agreement was 93.3\%, with Cohen’s $\kappa = 0.865$ (Appendix Table~\ref{tab:aopsreliability}). Disagreements were then resolved through discussion, yielding a stabilized coding scheme that was applied to the remaining problems. In a small number of cases, subsequent model outputs revealed additional distinctions or equivalences that warranted refinement of the problem-specific strategy families. These revisions were reviewed and incorporated into the final taxonomy. 
\fref{fig:aopsexample} illustrates an example of the resulting strategy classification. Additional Examples can be found in Appendix~\ref{app:coding}.

\section{Experiments}
\subsection{Models and prompting protocol}
We evaluate four frontier reasoning models from major publicly available LLM ecosystems: GPT-5.4, Gemini-3.1-Pro, Claude Opus 4.6, and DeepSeek-V3.2 (Reasoner). Our goal is not merely to compare models, but to use a set of widely used frontier systems to characterize the current state of mathematical reasoning under the proposed evaluation framework. For each framework problem, every model is queried once per problem per prompt condition. In the \emph{baseline solve prompt}, the model is asked to solve the problem and provide its final answer. In the \emph{multiple-strategy prompt}, the model is asked to solve the same problem using multiple distinct strategies whenever possible. The structured JSON output includes \texttt{strategy\_name}, \texttt{method\_summary}, \texttt{key\_steps}, and \texttt{final\_answer} (Appendix~\ref{app:prompt}). Unless otherwise noted, all diversity analyses are based on the multiple-strategy condition, while the baseline condition is used primarily for conventional solve accuracy. This design is intended to capture realistic model behavior under a common task framing and structured-output protocol, generally using provider-default inference settings rather than fully budget-matched decoding controls. See \href{https://github.com/Emily2021Yang/math-strategy-diversity-eval.git}{detals} at Github.

\subsection{Coding pipeline and reliability}
\begin{figure}[tbp]
  \centering
  \includegraphics[width=0.92\columnwidth]{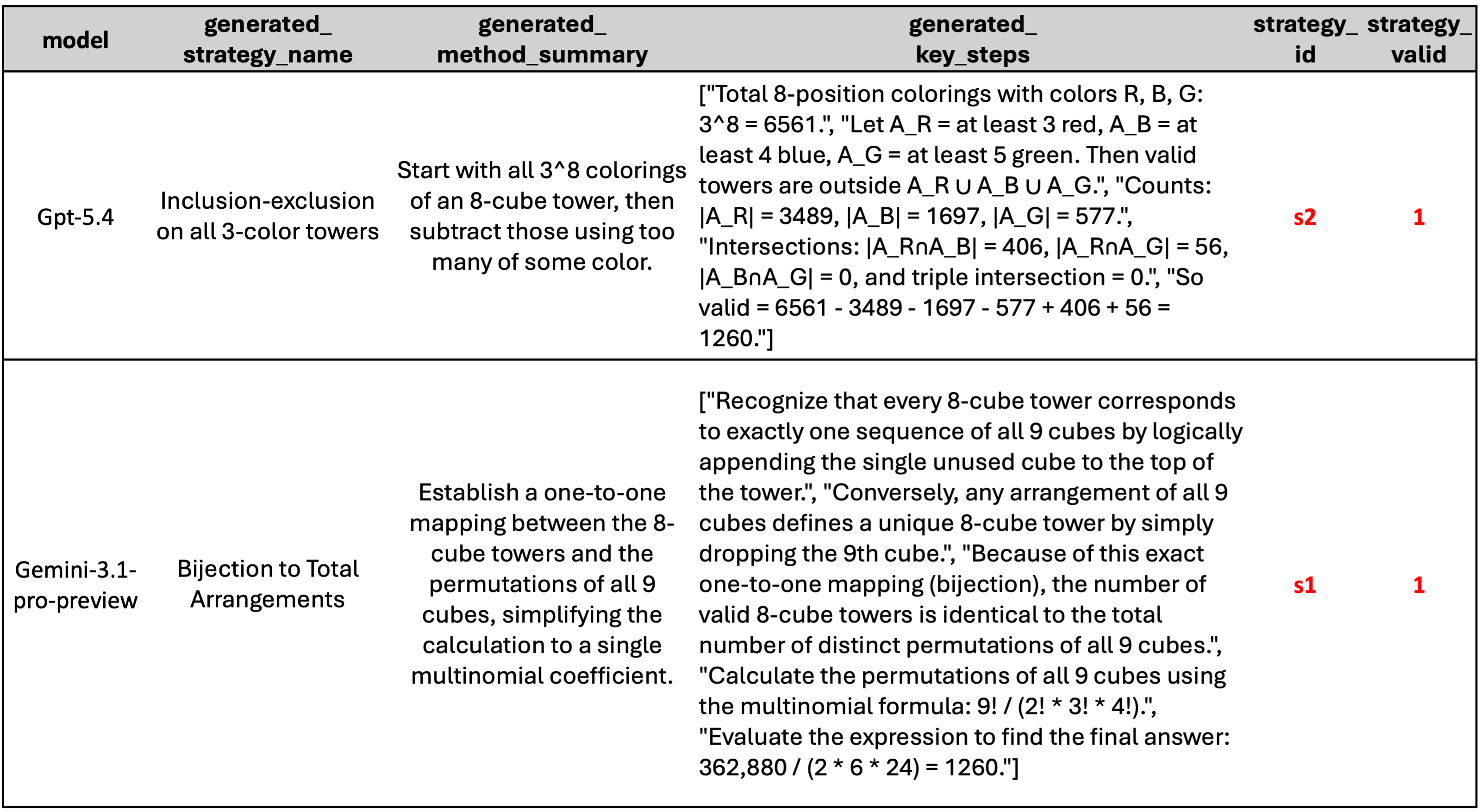}
\caption{Illustrative example of a model-generated solution record within the coding pipeline.}
\label{fig:outputexample}
\end{figure}
Model outputs are annotated using a hybrid pipeline that combines dual-AI coding with targeted human review. For each response, two independent AI coders, GPT-5.4 and Gemini-3.1-flash, are provided with the problem statement, the problem-specific reference strategies, and four in-context coding examples (Appendix~\ref{app:promptexample}). Each coder then assigns three labels: \texttt{strategy\_id}, \texttt{strategy\_valid} (0/1), and \texttt{result\_correct} (0/1). A \texttt{strategy\_id} is assigned by matching each output to the problem-specific reference strategy families. Outputs that align with the reference taxonomy are labeled $s_1, s_2, \ldots$, while those that do not match any reference family are initially labeled \emph{novel}. For outputs labeled \emph{novel}, those that are both valid and correct are further designated as novel families $n_1, n_2, \ldots$. Using the same problem and reference strategies shown in \fref{fig:aopsexample}, \fref{fig:outputexample} illustrates how a model output is coded.

Agreement between the two AI coders was strong for \texttt{strategy\_id} (\tref{tab:reliability}). Out of 924 coded outputs, 184 rows exhibited disagreement and were adjudicated by human coders. To assess the reliability of AI-coded labels, human coders independently reviewed sampled subsets of AI-agreement rows: 267 rows for \texttt{strategy\_valid} and 261 rows for \texttt{strategy\_id}. The smaller sample for \texttt{strategy\_id} reflects that invalid responses are not assigned a finalized strategy identifier. Human--AI reliability was high for both \texttt{strategy\_id} and \texttt{strategy\_valid} (\tref{tab:reliability}).

\begin{table}[H]
\centering
\caption{Coding reliability for strategy annotation. Human--AI reliability is computed on sampled AI-agreement rows. Sample size differ because \texttt{strategy\_id} applies only to valid strategies. $\kappa$ is not reported for \texttt{result\_correct}, which is directly verifiable.}
\label{tab:reliability}
\small
\begin{tabularx}{\columnwidth}{>{\raggedright\arraybackslash}p{0.22\columnwidth}>{\raggedright\arraybackslash}p{0.23\columnwidth}>{\raggedright\arraybackslash}p{0.17\columnwidth}cc}
\toprule
Comparison & Subset & Field & $N$ & $\kappa$ / Agreement \\
\midrule
\multirow{3}{*}{GPT vs Gemini} & \multirow{3}{*}{All coded rows} & strategy\_id & 924 & 0.805 / 85.4\% \\
& & strategy\_valid & 924 & 0.651 / 93.9\% \\
& & result\_correct & 924 & 0.950 / 99.6\% \\
\midrule
\multirow{2}{*}{Human vs agreed AI} & \multirow{2}{*}{\makecell[l]{Sampled\\AI-agreement rows}} & strategy\_id & 261 & 0.793 / 84.7\% \\
& & strategy\_valid & 267 & 0.958 / 99.6\% \\
\bottomrule
\end{tabularx}
\end{table}

\subsection{Evaluation metrics}
We evaluate models using a suite of metrics derived from the finalized strategy annotations. For context, we also report final-answer accuracy. Our framework-specific metrics fall into three groups: (i) \textbf{strategy diversity} measures the breadth of valid solution families generated by each model, using both the total number of valid distinct strategies and the average number of distinct strategies per problem. Confidence intervals for paired gaps are estimated by bootstrap resampling over problems, and domain-level gaps are also reported. (ii) \textbf{AoPS alignment} compares model-generated strategies with the human reference set, using AoPS coverage and the mean paired diversity gap relative to AoPS across matched problems. AoPS coverage is defined as the proportion of reference strategies recovered by a model; (iii) \textbf{AoPS-novel strategies} capture valid model-generated strategies not represented in the AoPS reference taxonomy, summarized by total count and by their distribution across domains and models.

\subsection{Repeated runs on a 20-problem subset}
We additionally study repeated sampling on a 20-problem subset to test whether the observed strategy inventory is robust to rerunning the same prompting-and-coding pipeline. The subset is intentionally balanced across domains, with four problems from each of Algebra, Combinatorics, Geometry, Number Theory, and Probability. This design keeps the manual auditing burden manageable while preserving domain coverage. We compare the original full-framework slice restricted to those 20 problems against two additional repeated runs, measuring both the union of discovered strategies and the incremental gains contributed by each rerun.

\section{Results}
\subsection{Prompt-level correctness}
We begin with final-answer correctness under both prompting conditions. As shown in \tref{tab:overalldiv}, all four models perform strongly on the 80-problem framework dataset under \texttt{prompt\_single}: Gemini achieves perfect accuracy, while DeepSeek, GPT, and Claude all exceed 95\%. Under \texttt{prompt\_multi}, correctness remains perfect for Gemini, improves slightly for GPT, declines modestly for DeepSeek, and drops more substantially for Claude. Thus, asking for multiple strategies does not uniformly preserve or improve solve success across models. At the model level, paired McNemar tests show that prompt effects are not statistically distinguishable for GPT, Gemini, or DeepSeek, whereas Claude has 8 problems solved only under \texttt{prompt\_single} but only 1 solved only under \texttt{prompt\_multi} (exact McNemar $p = 0.039$; Appendix Table~\ref{tab:appendixmcnemar}). Domain-level changes under \texttt{prompt\_multi} are limited overall, but Claude shows a notable decline in Geometry, from 100.0\% to 75.0\%. This 25-point drop corresponds to four geometry problems (Appendix Table~\ref{tab:appendixinventory}).

\begin{table}[H]
\centering
\caption{Final-answer correctness by prompt. For \texttt{prompt\_single}, accuracy is computed over all 80 framework problems. For \texttt{prompt\_multi}, a problem counts as correct if at least one generated strategy is labeled \texttt{result\_correct = 1}.}
\label{tab:overalldiv}
\small
\setlength{\tabcolsep}{2.5pt}
\renewcommand{\arraystretch}{0.96}
\begin{tabular}{lccccc}
\toprule
& \multicolumn{2}{c}{\texttt{prompt\_single}} & \multicolumn{3}{c}{\texttt{prompt\_multi}} \\
\cmidrule(lr){2-3} \cmidrule(lr){4-6}
Source & Correct / Total & Acc. & Correct / Total & Acc. & $\Delta$ Acc. \\
\midrule
GPT & 77 / 80 & 96.2\% & 78 / 80 & 97.5\% & +1.3 \\
Gemini & 80 / 80 & 100.0\% & 80 / 80 & 100.0\% & 0.0 \\
DeepSeek & 79 / 80 & 98.8\% & 77 / 80 & 96.2\% & -2.6 \\
Claude & 76 / 80 & 95.0\% & 69 / 80 & 86.2\% & -8.8 \\
\bottomrule
\end{tabular}
\renewcommand{\arraystretch}{1}
\end{table}

\subsection{Overall strategy diversity and AoPS alignment}
\begin{figure}[H]
  \centering
  \includegraphics[width=0.40\columnwidth]{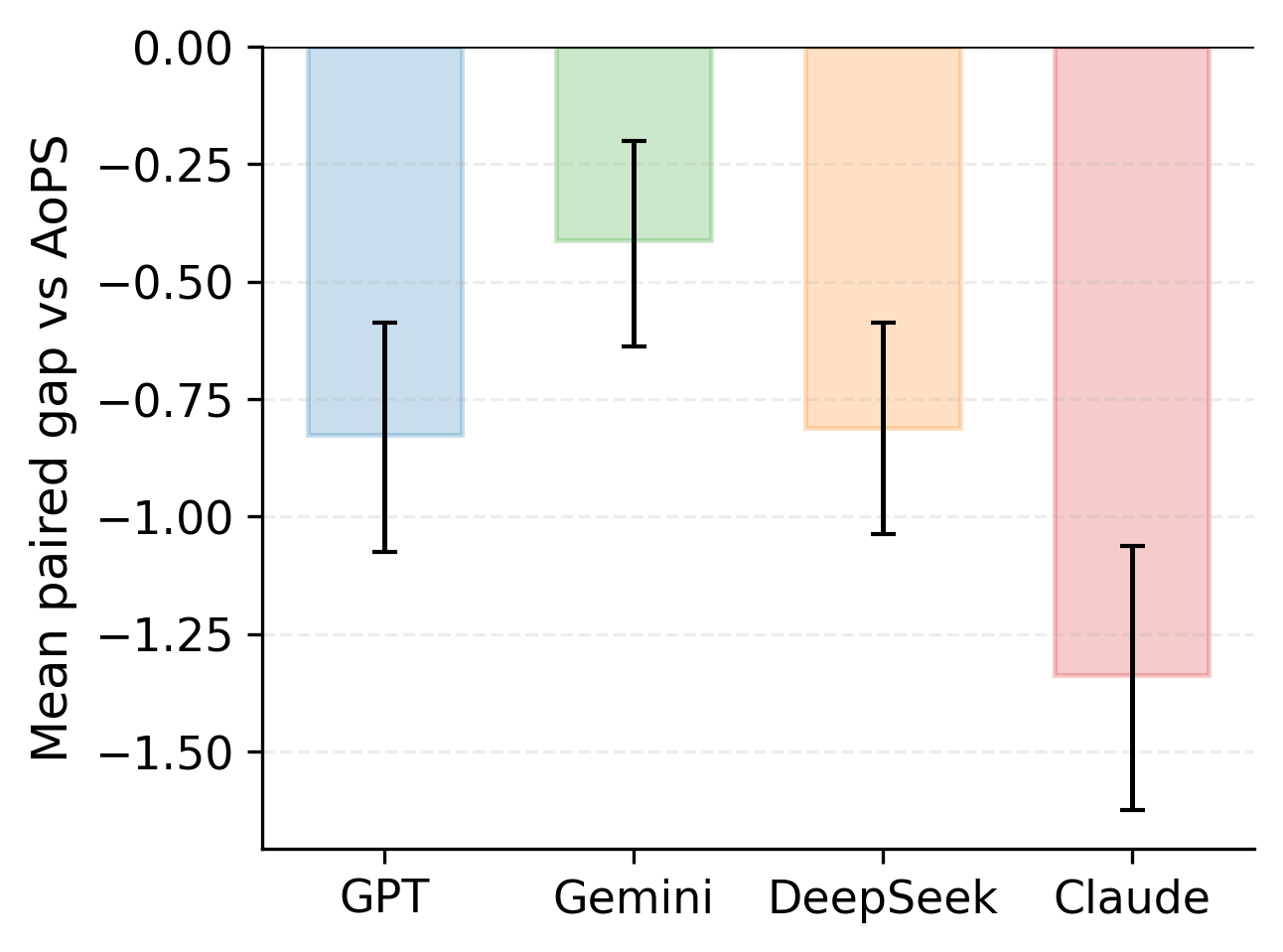}
  \caption{Overall paired strategy-diversity gap relative to AoPS across 80 problems. Bars show the mean paired gap for each model, and error bars denote 95\% bootstrap confidence intervals. Negative values indicate fewer distinct strategies than the AoPS reference on matched problems.}
  \label{fig:overallgap}
\end{figure}

Across the 80 problems, the AoPS corpus contains 217 distinct strategies (2.71 strategies per problem on average), establishing the reference inventory for strategy diversity. All four models fall below this benchmark in total distinct strategies. Gemini comes closest, producing 184 distinct strategies (2.30 per problem), followed by DeepSeek (152), GPT (151), and Claude (110). Total strategy count, however, do not directly correspond to AoPS-reference coverage, as part of each model's inventory consists of strategies not present in the AoPS corpus. Relative to the 217-strategy AoPS reference pool, Gemini recovers 150 strategies (69.1\%), GPT 130 (59.9\%), DeepSeek 128 (59.0\%), and Claude 100 (46.1\%) (\tref{tab:domaincov}). \fref{fig:overallgap} summarizes the mean paired diversity gaps and 95\% bootstrap confidence intervals using the full 80-problem comparison. All mean gaps are negative, indicating systematic underperformance relative to the human reference even after matching models and AoPS on the same problems (Appendix Table~\ref{tab:appendixgap}).

\begin{table}[H]
\centering
\caption{Coverage of the AoPS reference pool and novel strategy generation by domain. ``Union Cov.'' denotes the AoPS-reference strategies recovered by the union of the four models. Model-specific coverage columns report each model's recovered AoPS-reference strategies relative to the fixed AoPS total in that domain.}
\label{tab:domaincov}
\scriptsize
\setlength{\tabcolsep}{2.5pt}
\resizebox{\columnwidth}{!}{%
\begin{tabular}{lrrrrrrrrr}
\toprule
Group & \#P & AoPS & Union Cov. & GPT Cov. & Gemini Cov. & DeepSeek Cov. & Claude Cov. & Novel & Novel/P \\
\midrule
Algebra & 14 & 37 & 36/37 (97.3\%) & 25/37 (67.6\%) & 29/37 (78.4\%) & 25/37 (67.6\%) & 27/37 (73.0\%) & 3 & 0.21 \\
Combinatorics & 19 & 43 & 32/43 (74.4\%) & 21/43 (48.8\%) & 25/43 (58.1\%) & 23/43 (53.5\%) & 17/43 (39.5\%) & 20 & 1.05 \\
Geometry & 16 & 53 & 38/53 (71.7\%) & 28/53 (52.8\%) & 34/53 (64.2\%) & 26/53 (49.1\%) & 16/53 (30.2\%) & 7 & 0.44 \\
Number Theory & 15 & 45 & 37/45 (82.2\%) & 28/45 (62.2\%) & 31/45 (68.9\%) & 29/45 (64.4\%) & 19/45 (42.2\%) & 4 & 0.27 \\
Probability & 16 & 39 & 32/39 (82.1\%) & 28/39 (71.8\%) & 31/39 (79.5\%) & 25/39 (64.1\%) & 21/39 (53.8\%) & 16 & 1.00 \\
\midrule
\textbf{All} & \textbf{80} & \textbf{217} & \textbf{175/217 (80.6\%)} & \textbf{130/217 (59.9\%)} & \textbf{150/217 (69.1\%)} & \textbf{128/217 (59.0\%)} & \textbf{100/217 (46.1\%)} & \textbf{50} & \textbf{0.63} \\
\bottomrule
\end{tabular}
\par}
\end{table}

\subsection{Domain-level strategy diversity and AoPS alignment}
The domain-level analyses are intended to be descriptive and comparative rather than absolute: the five domains differ in their typical strategy multiplicity and problem structure, and the number of problems per domain is modest. Accordingly, the goal is to identify where the model-AoPS gap appears relatively larger or smaller within this framework, not to claim that raw diversity counts are directly exchangeable across domains. 

\begin{figure}[H]
  \centering
  \begin{minipage}[t]{0.48\columnwidth}
    \centering
    \includegraphics[width=\linewidth]{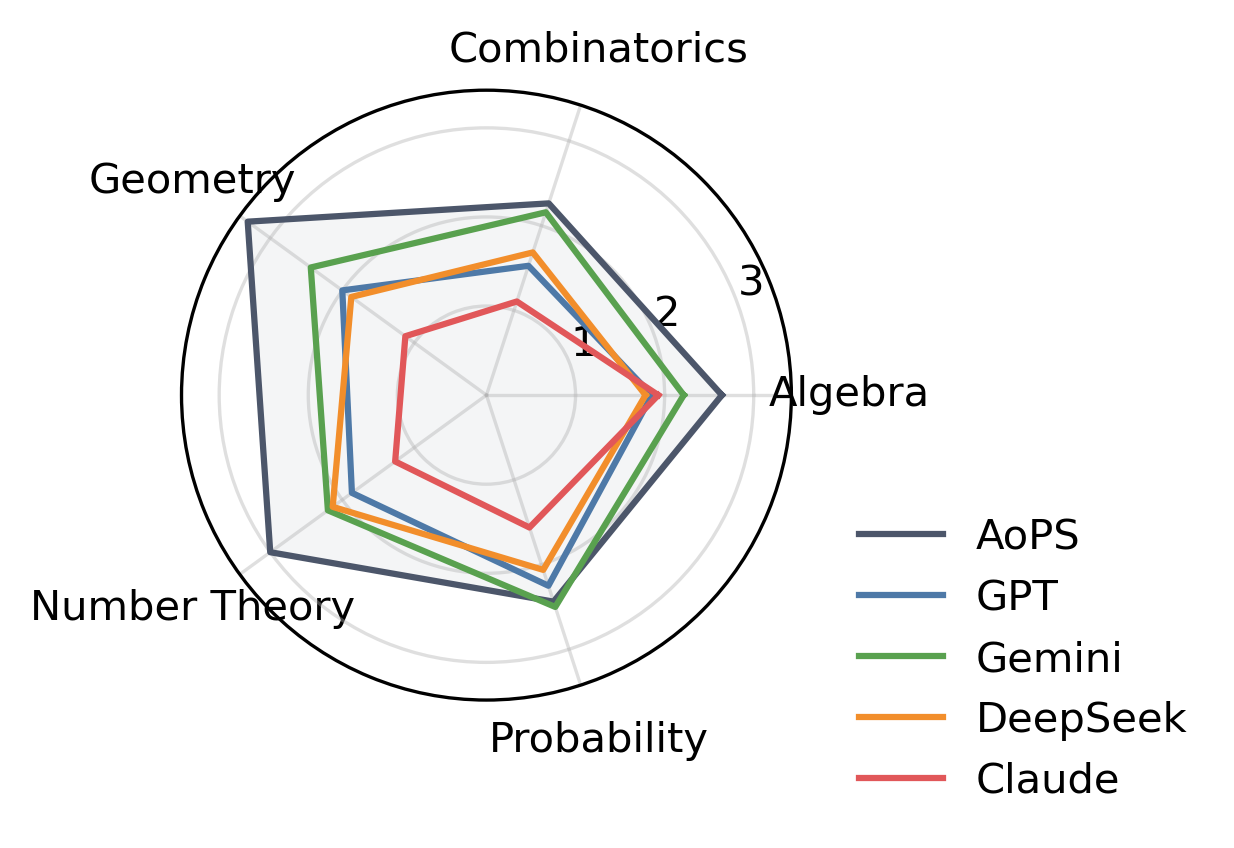}
    \captionof{figure}{Radar view of domain-level strategy diversity. AoPS exhibits the richest strategy diversity overall, with the strongest human advantage concentrated in Geometry and Number Theory.}
    \label{fig:radar}
  \end{minipage}\hfill
  \begin{minipage}[t]{0.5\columnwidth}
    \centering
    \includegraphics[width=\linewidth]{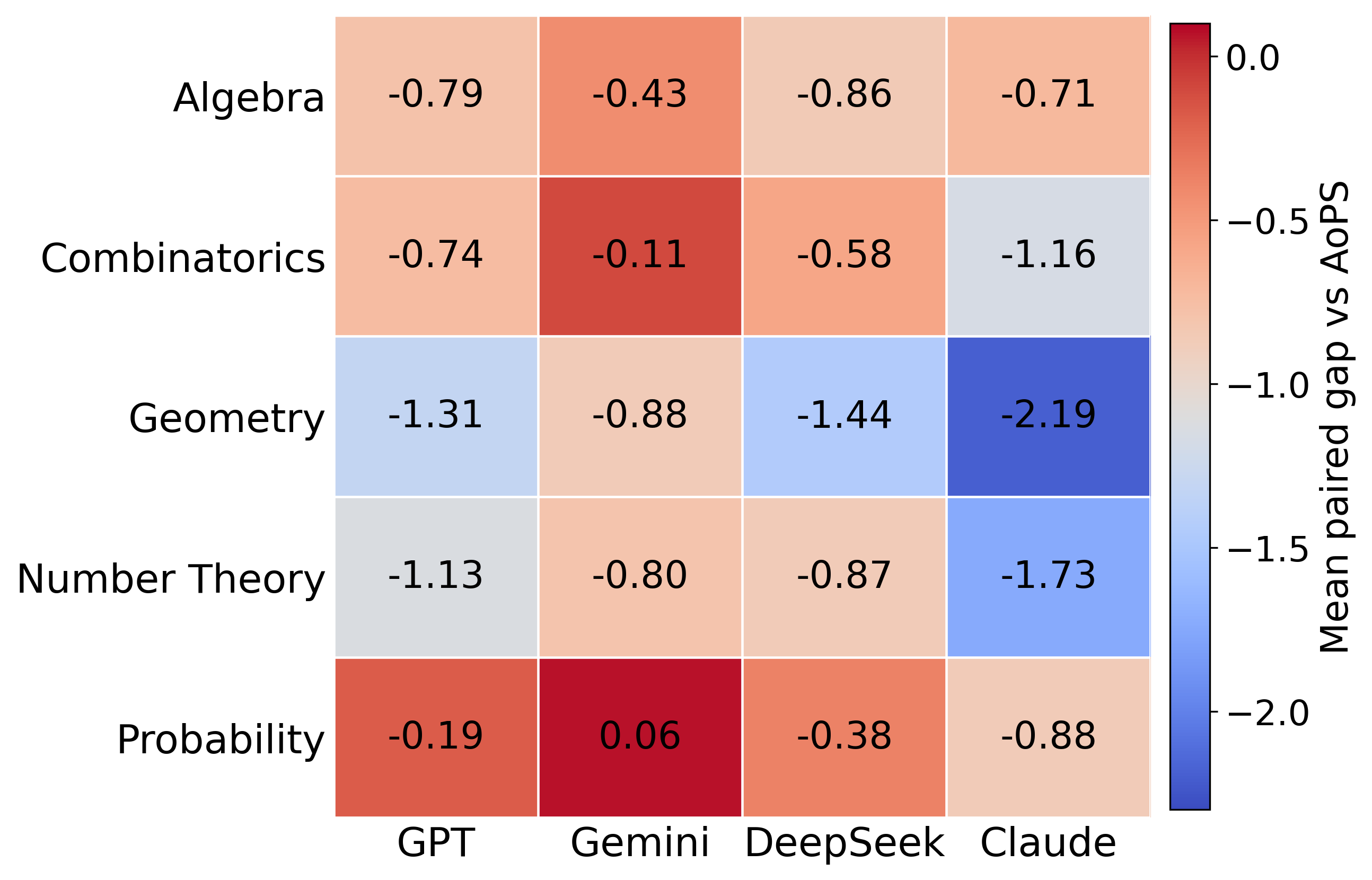}
\captionof{figure}{Domain-level strategy-diversity gap relative to AoPS. Each cell shows the mean paired gap for one model-domain combination; more negative values indicate fewer distinct strategies than the AoPS reference on matched problems.}
    \label{fig:domaingap}
\end{minipage}
\end{figure}

\fref{fig:radar} provides a descriptive domain-level view of average strategy diversity, while \fref{fig:domaingap} summarizes paired gaps relative to AoPS.  First, AoPS reference shows the highest average diversity in Geometry and Number Theory, suggesting that these domains support especially rich spaces of alternative solution methods. Second, model deficits are likewise domain-dependent. The largest gaps appear in Geometry, where all models fall well below the human reference and Claude shows the strongest deficit. Substantial gaps also appear in Number Theory. By contrast, smaller deficits are observed in Algebra, Combinatorics, and especially Probability. In Probability, Gemini comes closest to the AoPS reference, and GPT also shows a relatively small gap. Overall, the domain pattern suggests that the human advantage in strategy diversity is not uniform across mathematics and is most pronounced in Geometry.

Next, we examine how AoPS-reference coverage vary across domains using both the union of the four models and individual models (\tref{tab:domaincov}). At the union-level, Algebra exhibits the highest AoPS coverage (97.3\%), whereas Geometry shows the weakest coverage (71.7\%). In terms of model-specific performance, Gemini outperforms the other models in every domain, with particularly high coverage in Probability and Algebra. Claude shows the weakest recovery overall and the largest shortfall in Geometry.

\subsection{AoPS-novel strategies generated by models}
\begin{figure}[H]
  \centering
  \includegraphics[width=0.68\columnwidth]{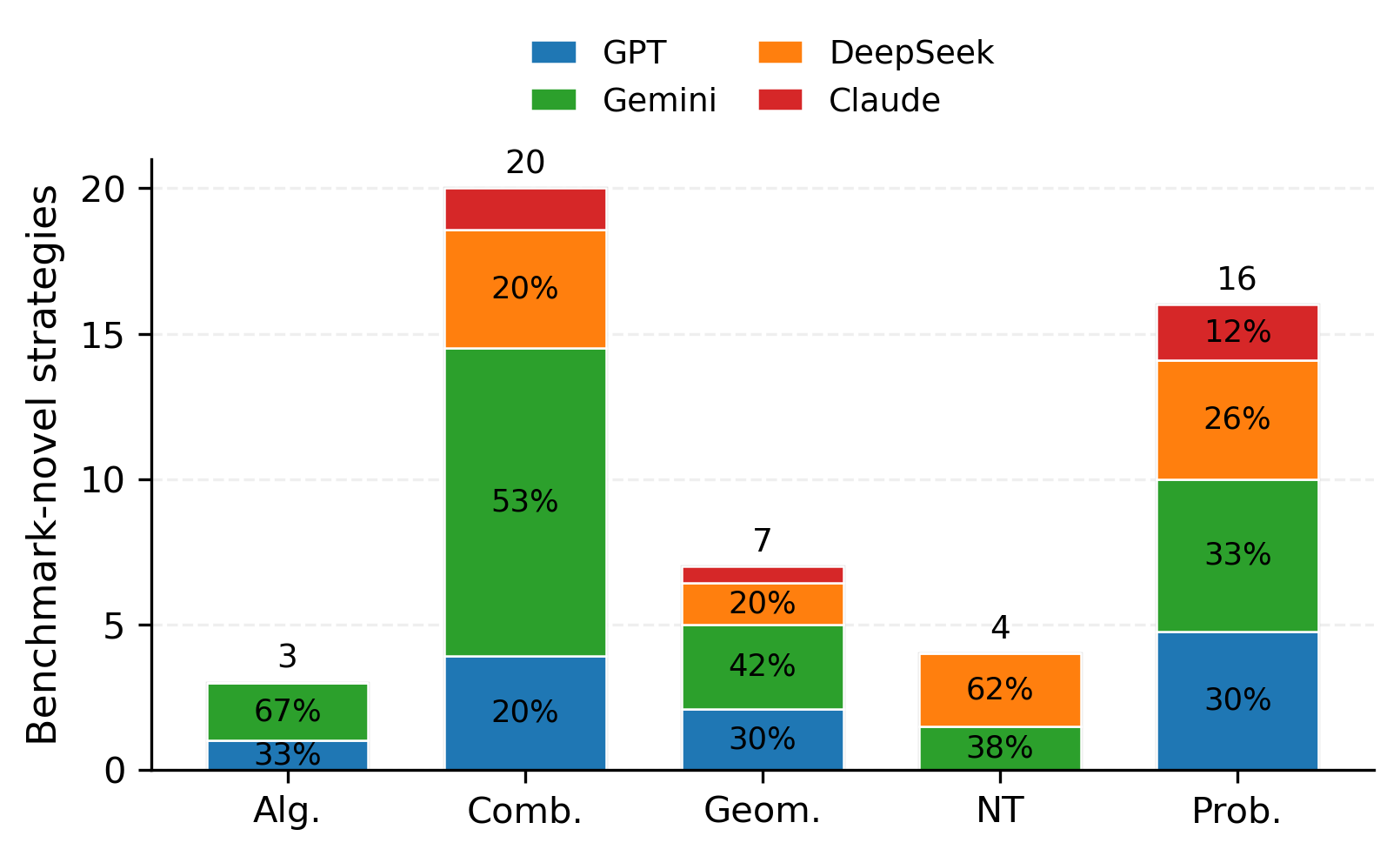}
\caption{Novel strategies by domain and model contribution. The height of each bar shows the total number of benchmark-novel valid strategies in the domain; colored segments indicate each model's fractional contribution, where a novel strategy shared by $k$ models gives each contributing model $1/k$ credit.}
  \label{fig:noveldomain}
\end{figure}

Across the 80-problem dataset, the four models collectively produce 50 novel valid strategies beyond the AoPS reference set. These counts are not mutually exclusive across models, as the same novel strategy may be independently discovered by multiple systems. Figure~\ref{fig:noveldomain} decomposes these totals by model and domain. Combinatorics is dominated by Gemini, Probability is more evenly shared across models, and the limited novelty in Number Theory is driven primarily by DeepSeek. Moreover, novel strategies are model-specific, while a smaller shared core is independently discovered by multiple systems (Appendix \fref{fig:upset}).

\subsection{Repeated-Run Robustness and Saturation}
The repeated-run of the 20-problem subset provides a targeted check of repeated-sampling behavior, but it should not be interpreted as a statistically representative stand-in for the full dataset or as a definitive estimate of full-framework saturation. Across all models, repeated sampling exhibits diminishing returns: the union of distinct strategies rises from 144 in the original slice to 167 after the first rerun and 177 after the second, with only two additional novel strategies discovered beyond the original slice.

\begin{figure}[H]
  \centering
  \includegraphics[width=\linewidth]{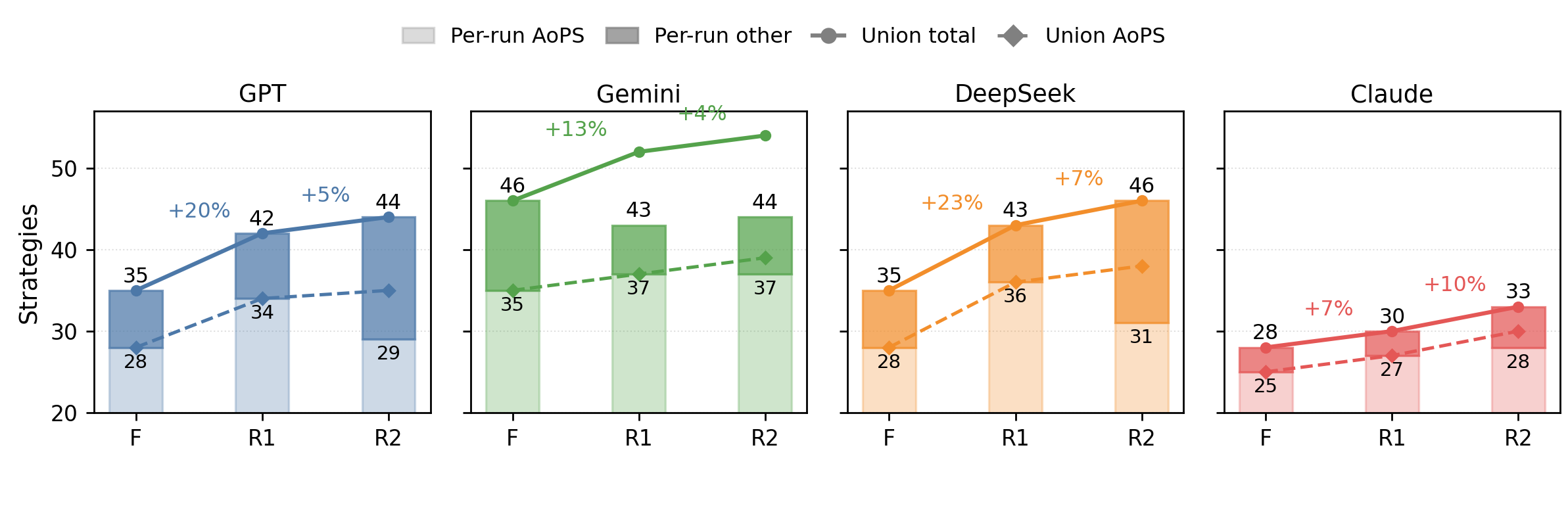}
\caption{Repeated-run strategy recovery on the balanced 20-problem subset. \texttt{F} is the original full-framework slice, \texttt{R1} the first rerun, and \texttt{R2} the second rerun. Stacked bars show per-run distinct strategies split into AoPS-reference and non-AoPS strategies. The solid line shows the cumulative union of total strategies, and the dashed line shows the cumulative union of AoPS-reference strategies. Percentage labels mark growth in the cumulative total-strategy union.}
  \label{fig:repeatpanels}
\end{figure}

At the model level, gains are uneven. DeepSeek and GPT show the largest first-rerun improvements in AoPS-reference recovery, increasing from 28 to 36 and from 28 to 34 recovered AoPS strategies, respectively; these increases are also the clearest statistically in the paired analysis (Appendix \tref{tab:repeatgainmodel}). Later gains are smaller and generally less statistically sharp across models. AoPS-reference coverage also grows more slowly than total strategy count. After all three runs, the strongest single-model recovery remains Gemini at 39/55 AoPS-reference strategies (70.9\%), followed by DeepSeek at 38/55 (69.1\%), GPT at 35/55 (63.6\%), and Claude at 30/55 (54.5\%), indicating a persistent gap to the human reference set.

\section{Conclusion}
We introduced a strategy diversity evaluation framework for mathematical problem solving by LLMs. Rather than focusing only on final-answer correctness, the framework compares model outputs against a human-coded AoPS reference set and measures how many distinct valid strategies models recover or contribute. Our experiments reveal a clear gap between answer accuracy and strategic flexibility: although all four evaluated models achieve strong final-answer performance, each falls short of the human reference set in distinct valid strategies generated. This gap is especially pronounced in Geometry and Number Theory. At the same time, models do contribute valid novel strategies, but these are concentrated disproportionately in Combinatorics and Probability. A repeated-run robustness check on a balanced 20-problem subset shows that additional sampling expands the discovered strategy inventory, but with diminishing marginal gains and without eliminating the gap to the AoPS reference set. These findings suggest that strategy diversity is an important lens for evaluating mathematical reasoning systems beyond answer accuracy alone.

\section{Limitations}
Because AMC/AIME problems and related solution materials are publicly available online, some benchmark items or canonical solution patterns may have appeared in model pretraining data or related web corpora \citep{mahdavi2025contamination}. Such exposure could inflate both final-answer accuracy and recovery of common AoPS-like strategies. We therefore interpret the framework as a comparative evaluation of strategy diversity against a human reference corpus, rather than as a contamination-free test of de novo mathematical reasoning. First, the current framework covers 80 competition-style problems, and repeated-run saturation was evaluated on only a balanced 20-problem subset. These scope choices reflect the substantial expert effort required to construct problem-specific strategy families and audit model-generated strategies, so the repeated-run findings should be interpreted as a targeted robustness check rather than a definitive estimate of full-framework saturation. Second, the human reference set is derived from AoPS community solutions, which are rich but incomplete; thus, novel strategies are novel relative to the collected AoPS corpus, not necessarily universally novel in mathematics. Third, cross-model comparisons use a common task framing but not fully normalized sampling controls: models were queried once per problem per prompt condition, provider defaults were often retained, and token budgets were not perfectly matched across systems. The results therefore reflect frontier model behavior under realistic provider-default conditions rather than strictly budget-matched estimates of attainable strategy diversity. Finally, strategy annotation remains a structured abstraction despite overlapping human coding, reliability checks, dual-AI coding, and human adjudication.

Future work can expand the framework to larger and more diverse mathematics datasets, broaden the human reference pool beyond AoPS, introduce budget-matched sampling comparisons, and develop more efficient annotation procedures for strategy-family construction.

\bibliographystyle{unsrtnat}
\bibliography{NeurIPS2026bib}

\appendix

\section{Inter-rater reliability for AoPS strategy annotation}
\begin{table}[H]
\centering
\caption{Inter-rater reliability for AoPS strategy coding. Units are temporary problem-specific slots used for coder comparison and not aligned with finalized strategy identifiers.}
\label{tab:aopsreliability}
\small
\begin{tabular}{lrrrr}
\toprule
Coder Pair & Problems & Units & Agreement & Cohen's $\kappa$ \\
\midrule
Coder A vs.\ B & 14 & 84 & 89.3\% & 0.786 \\
Coder C vs.\ D & 16 & 96 & 96.9\% & 0.934 \\
Pooled & 30 & 180 & 93.3\% & 0.865 \\
\bottomrule
\end{tabular}
\end{table}

This appendix reports inter-rater reliability for the initial human coding used to align AoPS solutions into problem-specific strategy families. Two independent coder pairs performed overlapping coding on 30 problems: Coder A vs.\ Coder B on 14 problems, and Coder C vs.\ Coder D on 16 problems. Because the coders did not always use the same raw strategy names, reliability was not computed on those raw names directly. Instead, for each problem we reconciled the provisional coder labels into temporary within-problem comparison slots ($s_1 - s_2$) and then measured whether each coder marked a given temporary slot as present or absent.

These temporary comparison labels are \emph{not} the finalized strategy names used in the benchmark release. They exist only for the initial cross-coder reliability check. Finalized AoPS strategy identifiers were assigned later, after coder discussion and reconciliation, and are reported in the released strategy inventory.

The full long-format coder-alignment file used for this reliability analysis is included in the released dataset as \texttt{aops\_coding\_reliability\_long.csv}; the public release is hosted on \href{https://doi.org/10.34740/kaggle/ds/10271409}{Kaggle}.

Table~\ref{tab:aopsreliability} summarizes the resulting reliability estimates. The pooled result is the main quantity reported in the paper: across the two coder pairs, agreement on the temporary within-problem strategy slots was 93.3\%, with Cohen's $\kappa = 0.865$.

\section{Prompting protocol}
\label{app:prompt}
This appendix records the prompt templates used in the benchmark. We distinguish two types of prompting: (i) generation prompts used to elicit model solutions and (ii) coding prompts used to classify generated strategies against the AoPS reference taxonomy.

\subsection{Generation prompts}
The benchmark uses two generation settings. The baseline \texttt{prompt\_single} elicits a single solution record, while \texttt{prompt\_multi} asks the model to provide multiple distinct strategies. The exact templates are:

\paragraph{\texttt{prompt\_single}}
\begin{quote}\ttfamily\small
Solve the following problem.\par
\medskip
Provide your answer in the structured format below. Keep explanations concise.\par
\medskip
Format:\par
\{\par
\ \ ``final\_answer'': ``...'',\par
\ \ ``method\_summary'': ``...'',\par
\ \ ``key\_steps'': [``...'', ``...'', ``...'']\par
\}\par
\medskip
Problem:\par
\{problem\}
\end{quote}

\paragraph{\texttt{prompt\_multi}}
\begin{quote}\ttfamily\small
Solve the following problem using MULTIPLE DISTINCT strategies.\par
\medskip
A distinct solution strategy is defined by the critical reasoning that determines the structural form of the solution and enables the problem to be solved.\par
Critical reasoning refers to the key conceptual or representational move that transforms the problem into a solvable form.\par
Differences in algebraic manipulation, computational order, or explanatory detail are not treated as distinct strategies if they rely on the same underlying reasoning structure.\par
\medskip
Keep each strategy concise.\par
\medskip
Format:\par
\{\par
\ \ ``strategies'': [\par
\ \ \ \ \{\par
\ \ \ \ \ \ ``strategy\_name'': ``...'',\par
\ \ \ \ \ \ ``method\_summary'': ``...'',\par
\ \ \ \ \ \ ``key\_steps'': [``...'', ``...'', ``...''],\par
\ \ \ \ \ \ ``final\_answer'': ``...''\par
\ \ \ \ \}\par
\ \ ]\par
\}\par
\medskip
Problem:\par
\{problem\}
\end{quote}

\subsection{Coding Prompt with in-context examples}
\label{app:promptexample}
Model-output coding was also prompt-based. Each AI coder received (i) the problem statement, (ii) the problem-specific AoPS reference strategies, (iii) the generated strategy to be coded, and (iv) four in-context hand-coded examples illustrating the intended decision rule. The coding prompt instructed the model to judge mathematical validity, compare the core method with the reference strategies for the same problem, assign exactly one provisional strategy label, and return structured JSON fields for strategy match, validity, and result correctness.

\paragraph{Coding prompt template.}
\begin{quote}\ttfamily\small
You are an expert coder for a research study on mathematical problem-solving strategies.\par
\medskip
Your task is to code ONE generated strategy by comparing it with the human reference strategies for the SAME problem.\par
\medskip
Coding order:\par
1. First judge whether the generated strategy is mathematically valid.\par
2. Identify the generated strategy's core mathematical method or reasoning structure.\par
3. Compare that core method with each reference strategy for the same problem.\par
4. If a reference strategy has \textquotedblleft equivalent\_notes\textquotedblright, use those notes as guidance about alternate phrasings or closely related methods that should be coded as equivalent to that reference strategy.\par
5. Judge whether the generated strategy reaches the target result.\par
6. Assign exactly one \textquotedblleft assigned\_strategy\_id\textquotedblright: a reference strategy\_id, \textquotedblleft novel\textquotedblright, or \textquotedblleft uncertain\textquotedblright.\par
\medskip
Important coding rules:\par
\medskip
1. Strategy match\par
Assign \textquotedblleft assigned\_strategy\_id\textquotedblright\ as follows:\par
- Use a reference \textquotedblleft strategy\_id\textquotedblright\ if the generated strategy uses the same core mathematical method or reasoning structure as that reference strategy.\par
- Use \textquotedblleft novel\textquotedblright\ if the generated strategy is clearly different from every reference strategy.\par
- Use \textquotedblleft uncertain\textquotedblright\ only if the generated strategy is too vague, incomplete, or overlaps multiple reference strategies so that no confident assignment is possible.\par
- Do not assign a strategy\_id only because the final answer is the same. The core method must match, including equivalent methods described in \textquotedblleft equivalent\_notes\textquotedblright.\par
\medskip
2. Validity\par
Code \textquotedblleft strategy\_valid\textquotedblright\ separately from strategy match:\par
- Use \textquotedblleft 1\textquotedblright\ if the generated strategy is mathematically valid and leads to the correct final answer.\par
- Use \textquotedblleft 0\textquotedblright\ if the generated strategy contains a mathematical error, unsupported claim, contradiction, or wrong final answer.\par
- Use \textquotedblleft uncertain\textquotedblright\ if there is not enough information to judge validity.\par
\medskip
3. Target result correctness\par
Code \textquotedblleft result\_correct\textquotedblright\ separately from strategy validity:\par
- Use \textquotedblleft 1\textquotedblright\ if the generated strategy reaches the correct target result.\par
- Use \textquotedblleft 0\textquotedblright\ if the generated strategy reaches an incorrect target result.\par
- Use \textquotedblleft uncertain\textquotedblright\ if the generated strategy does not provide enough information to judge the target result.\par
- For most problems, the target result is the problem's final answer.\par
- If a coding item has \textquotedblleft coding\_scope\_note\textquotedblright, use that note to define the target result instead of the full final answer.\par
\medskip
4. Important distinction\par
A generated strategy can:\par
- match a reference strategy and be valid;\par
- match a reference strategy but be invalid;\par
- be novel and valid;\par
- be novel and invalid;\par
- be uncertain in match but still valid or invalid;\par
- have invalid reasoning but accidentally reach the correct target result;\par
- use valid reasoning but report an incorrect target result because of arithmetic or transcription error.\par
\medskip
5. Missing fields\par
Some generated strategies may have a blank \textquotedblleft generated\_method\_summary\textquotedblright.\par
If so, judge using the strategy name, key steps, final answer, problem text, and reference strategies.\par
\medskip
6. Confidence\par
Use confidence scores between 0 and 1:\par
- \textquotedblleft 1.0\textquotedblright\ means completely confident.\par
- \textquotedblleft 0.8\textquotedblright\ means fairly confident.\par
- \textquotedblleft 0.5\textquotedblright\ means uncertain or weak evidence.\par
- Below \textquotedblleft 0.5\textquotedblright\ should be rare.\par
\medskip
Return ONLY valid JSON with this exact schema:\par
\{\par
\ \ \textquotedblleft assigned\_strategy\_id\textquotedblright: \textquotedblleft ...\textquotedblright,\par
\ \ \textquotedblleft strategy\_match\_confidence\textquotedblright: 0.0,\par
\ \ \textquotedblleft strategy\_valid\textquotedblright: 0,\par
\ \ \textquotedblleft validity\_confidence\textquotedblright: 0.0,\par
\ \ \textquotedblleft result\_correct\textquotedblright: 0\par
\}\par
\medskip
Hand-coded examples:\par
\{\{examples\_json\}\}\par
\medskip
Coding item:\par
\{\{coding\_item\_json\}\}
\end{quote}

\paragraph{Representative in-context example.}
Table~\ref{tab:promptingexample} shows one of the four hand-coded demonstrations supplied to the AI coders. In this example, $s_1$ refers to the finalized problem-specific strategy identifier assigned in the AoPS taxonomy for this problem.

\begin{table}[H]
\centering
\caption{Representative hand-coded example used in the four-shot coding prompt. The full set of four in-context examples is included in the released materials as \texttt{coding\_examples\_nshot.csv}.}
\label{tab:promptingexample}
\small
\begin{tabularx}{\linewidth}{>{\raggedright\arraybackslash}p{0.24\linewidth}X}
\toprule
Field & Content \\
\midrule
Example ID & e5 \\
Problem ID & 2022AMC10A24 \\
Generated strategy name & Pollak's Cyclic Shift / Symmetry Argument \\
Generated method summary & Model the problem using a cyclic parking arrangement with an extra spot to map the threshold conditions into a uniform probabilistic space using rotational symmetry. \\
Generated key steps & Imagine 5 drivers selecting preferred spots from $\{0,1,2,3,4,5\}$ on a circular track with 6 spots; each driver attempts to park at the preferred spot and moves to the next available spot modulo 6 if needed; a sequence leaves spot 5 empty exactly when the original digit condition is satisfied; by rotational symmetry, one-sixth of the $6^5$ preference sequences leave spot 5 empty. \\
Generated final answer & 1296 \\
Assigned strategy ID & s1 \\
Strategy match confidence & 1 \\
Strategy valid & 1 \\
Validity confidence & 1 \\
Result correct & 1 \\
\bottomrule
\end{tabularx}
\end{table}

The full prompt template and all four in-context examples are included in the released code and dataset materials on \href{https://github.com/Emily2021Yang/math-strategy-diversity-eval.git}{GitHub} and \href{https://doi.org/10.34740/kaggle/ds/10271409}{Kaggle}.

\section{Coding examples}
\label{app:coding}
This section provides representative examples of the coding decisions used throughout the evaluation framework. The goal is to clarify how problem-specific strategy families were defined and how borderline cases were resolved. The examples cover five functions: distinguishing similar-looking but genuinely different strategies, separating invalid reasoning from correct final answers, validating benchmark-novel strategies, assigning a dominant critical step in multi-stage solutions, and resolving AI-coder disagreements through human adjudication.

\subsection{Example 1: Same-family vs different-family boundary case}
This example highlights that solutions with similar surface computations may belong to different strategy families when their dominant reasoning steps differ.

\noindent\includegraphics[width=\linewidth]{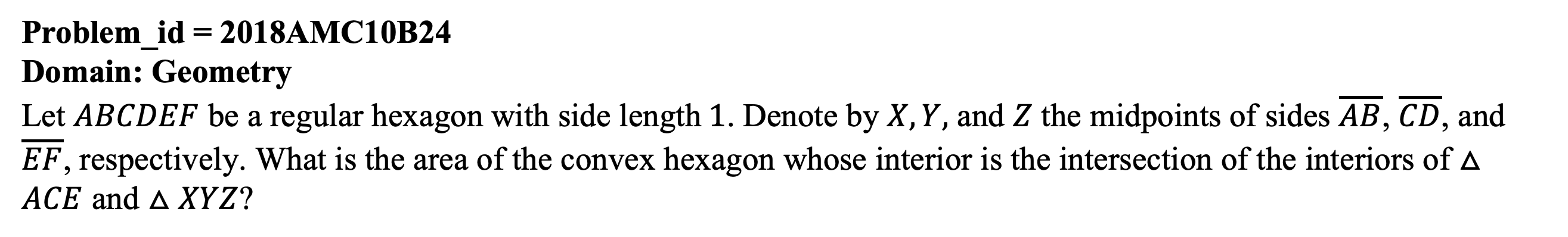}\par\vspace{2pt}
\noindent\includegraphics[width=\linewidth]{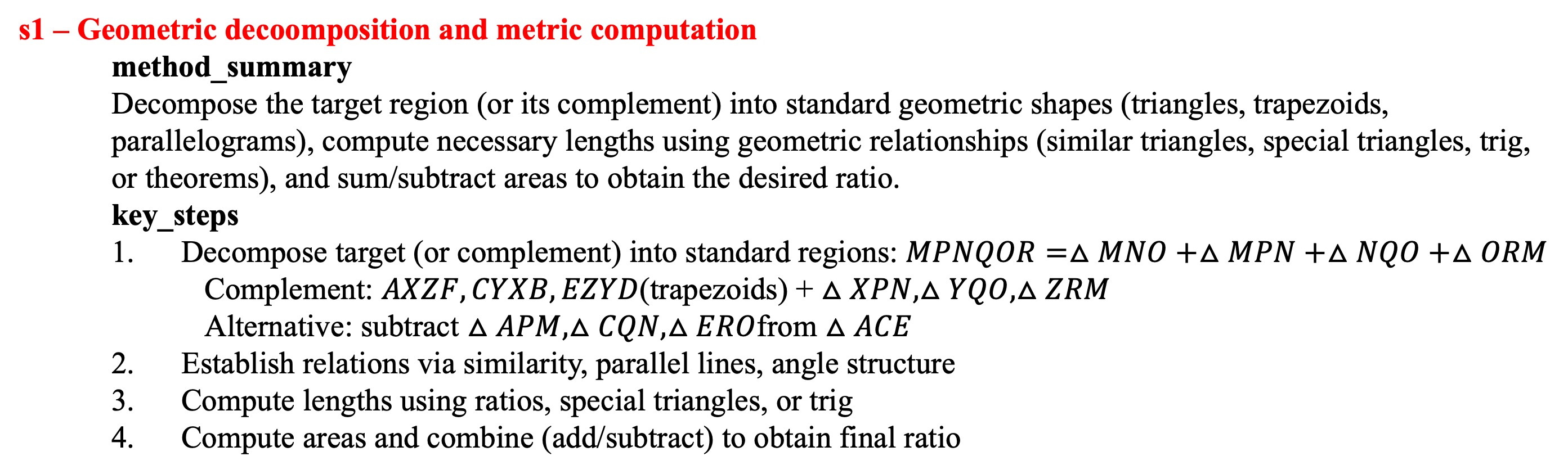}\par\vspace{2pt}
\noindent\includegraphics[width=\linewidth]{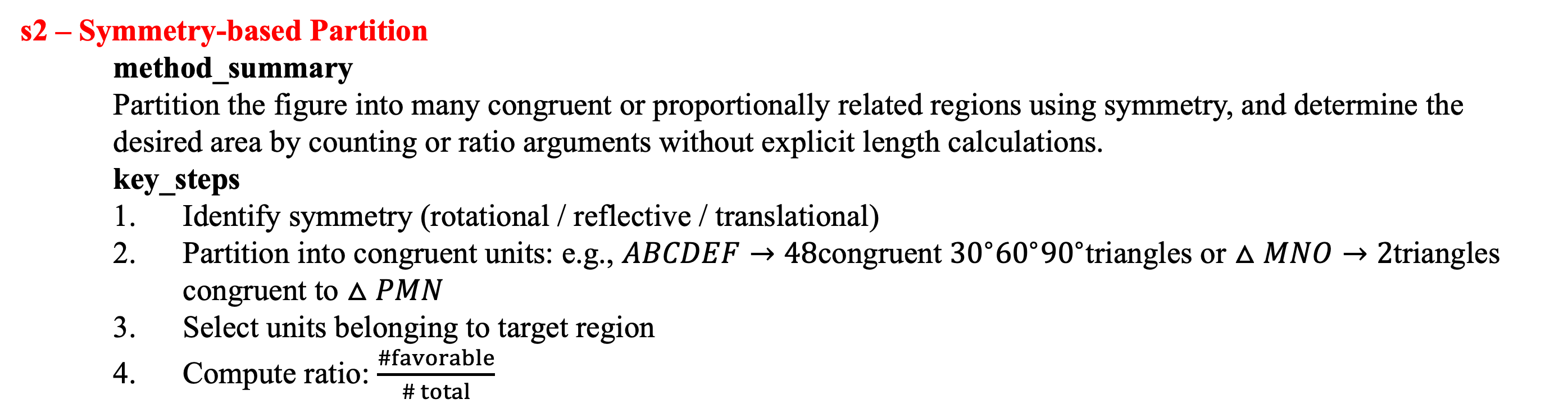}\par\vspace{2pt}
\noindent\includegraphics[width=\linewidth]{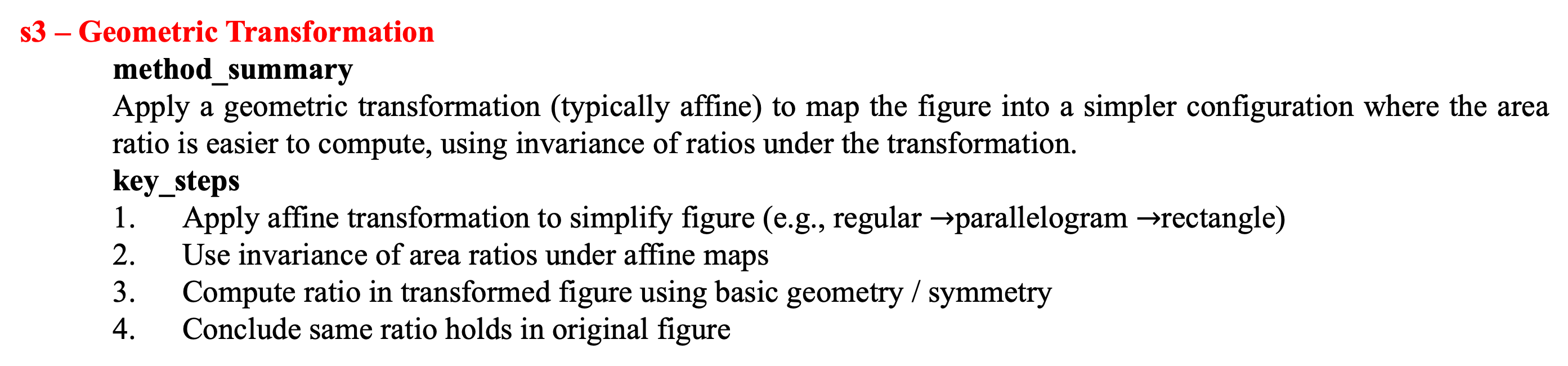}\par\vspace{2pt}
\noindent\includegraphics[width=\linewidth]{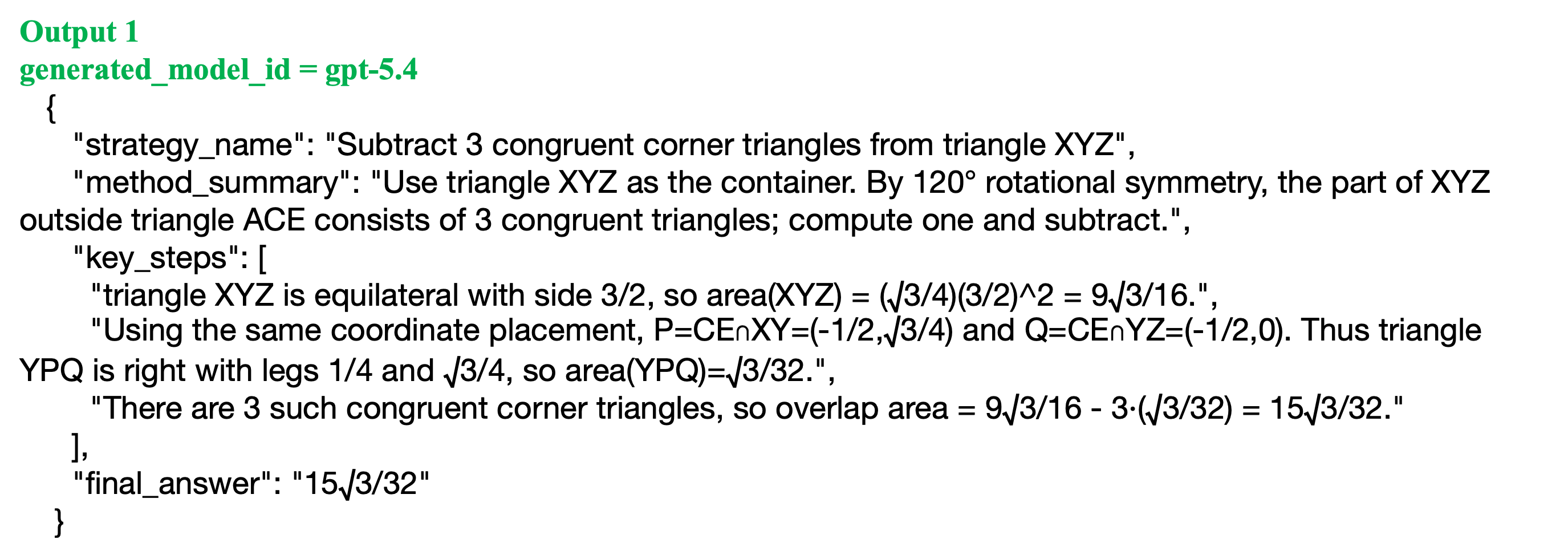}\par\vspace{2pt}
\noindent\includegraphics[width=\linewidth]{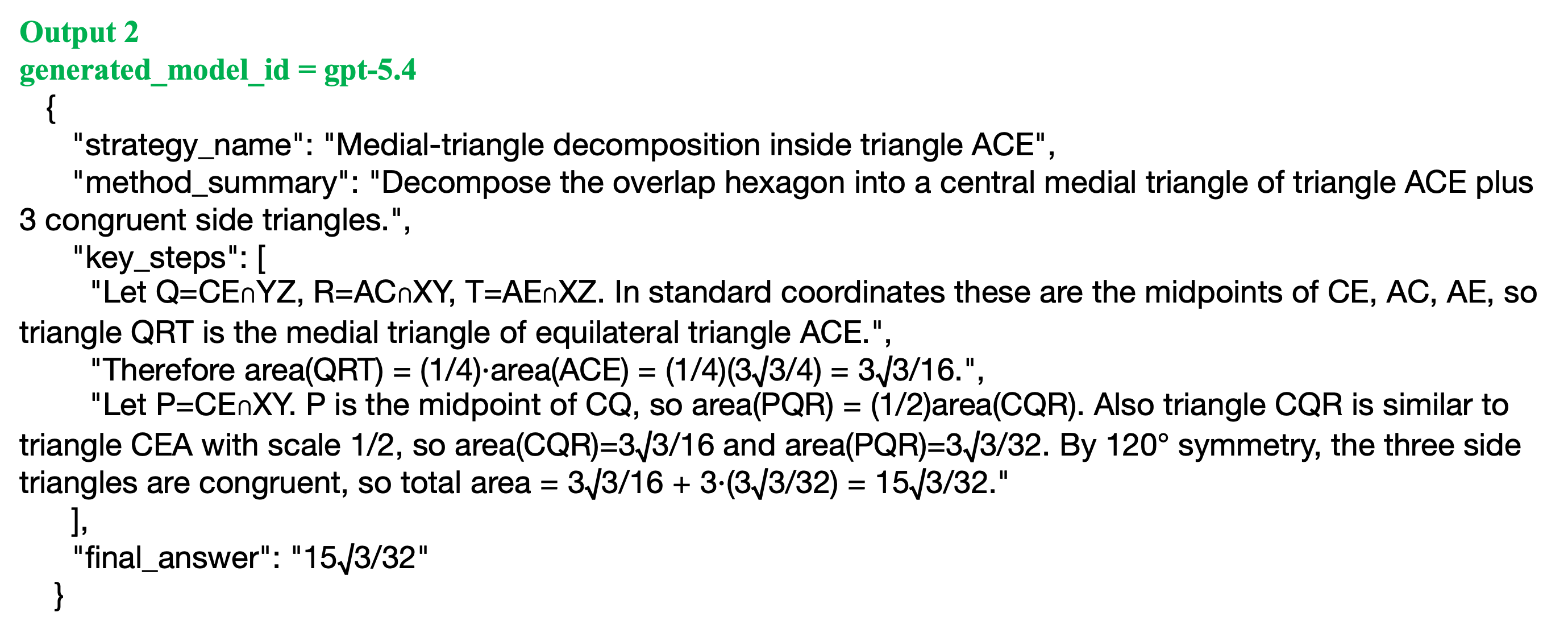}\par\vspace{2pt}
\noindent\includegraphics[width=\linewidth]{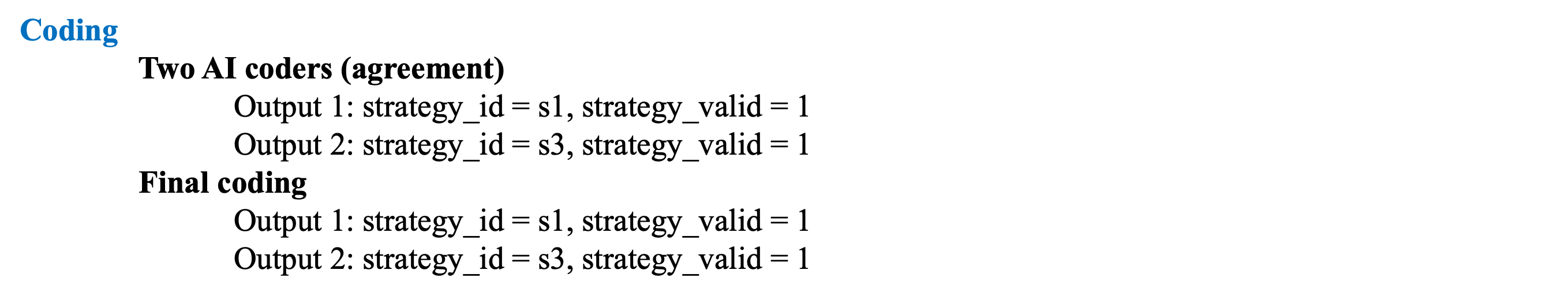}\par\vspace{2pt}

\newpage
\subsection{Example 2: Invalid strategy vs correct final answer}
This example uses the problem in Example 1 and shows why a correct final answer can still be coded as an invalid strategy.

\noindent\includegraphics[width=\linewidth]{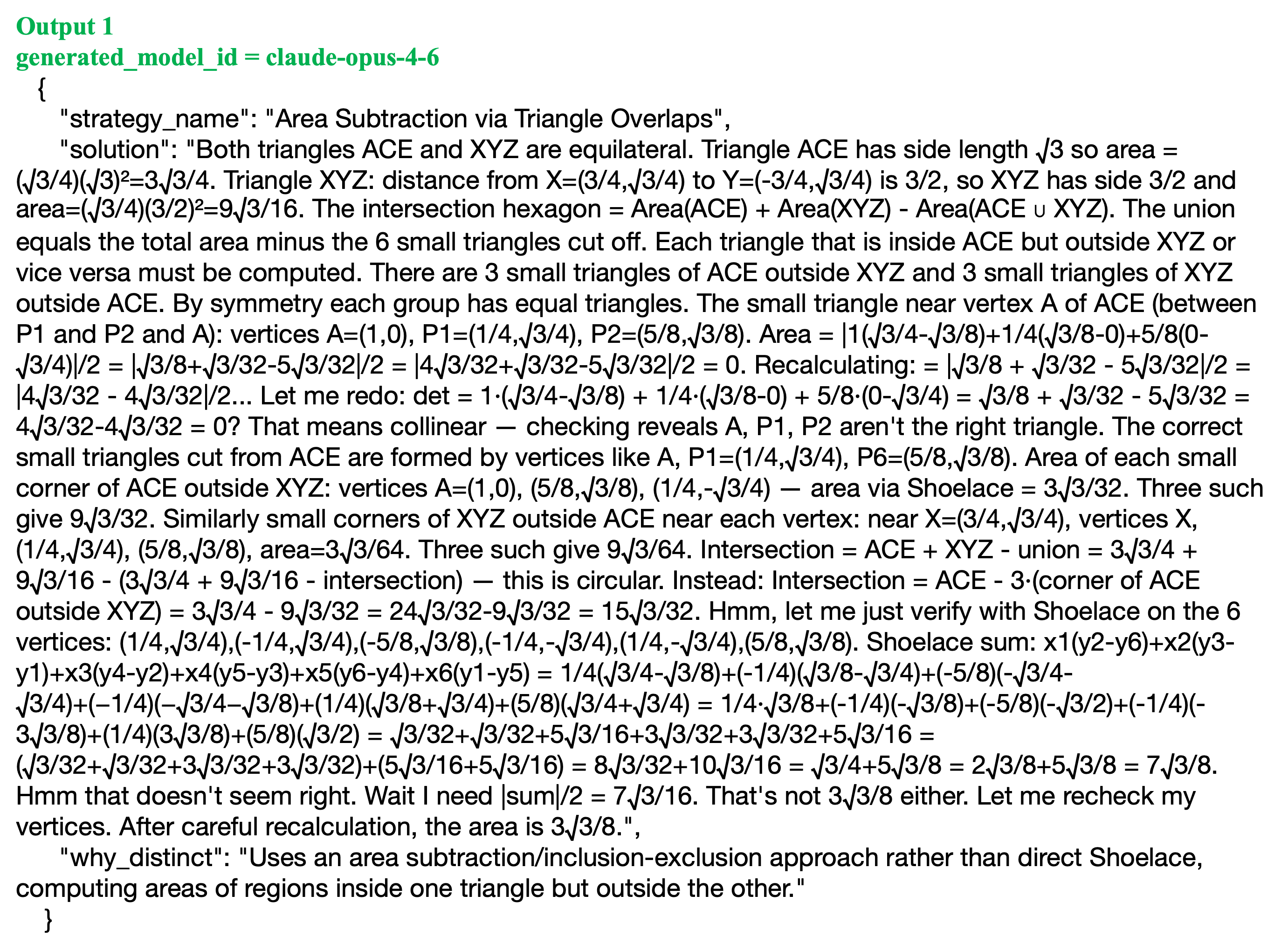}\par\vspace{2pt}
\noindent\includegraphics[width=\linewidth]{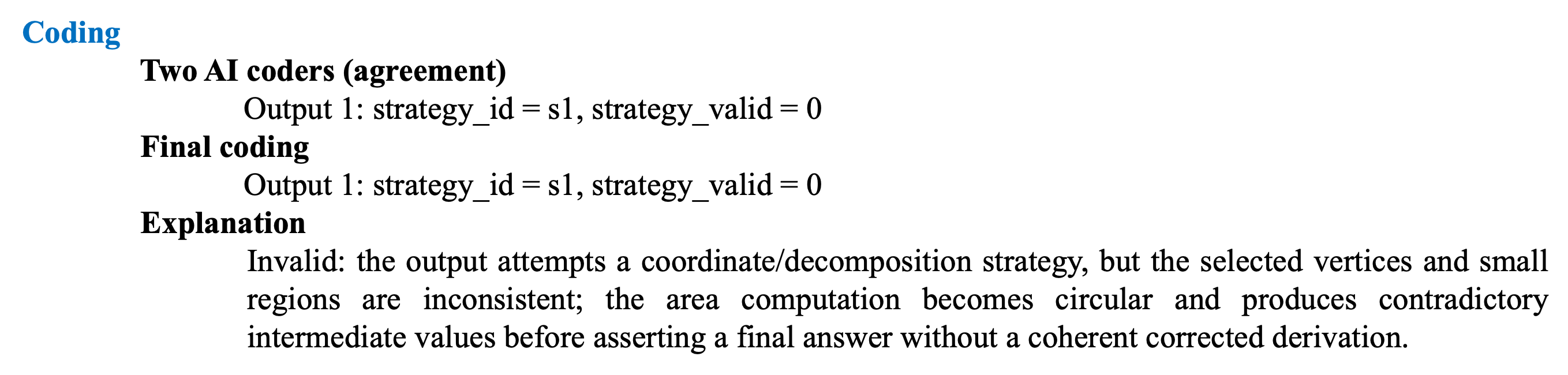}\par\vspace{2pt}

\subsection{Example 3: AoPS-novel valid strategy}
This example shows how strategies are labeled as novel when they employ core reasoning mechanisms absent from the AoPS reference set, with labels determined by the dominant critical step.
\noindent\includegraphics[width=\linewidth]{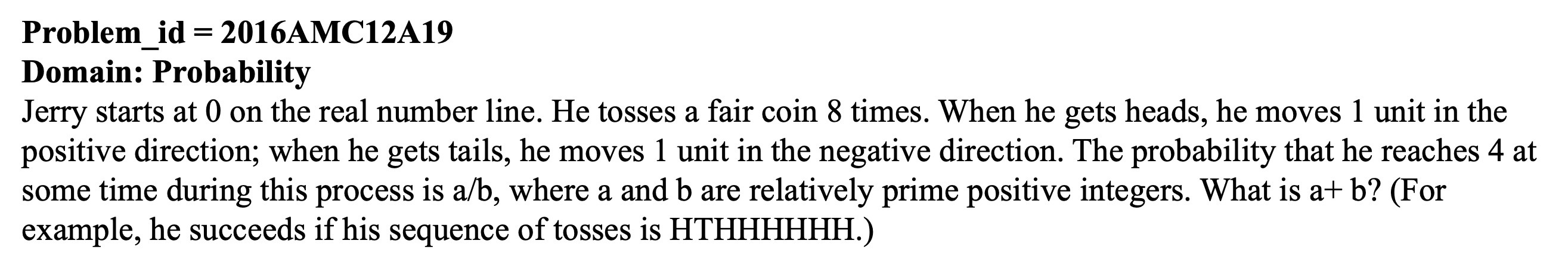}\par\vspace{2pt}
\noindent\includegraphics[width=\linewidth]{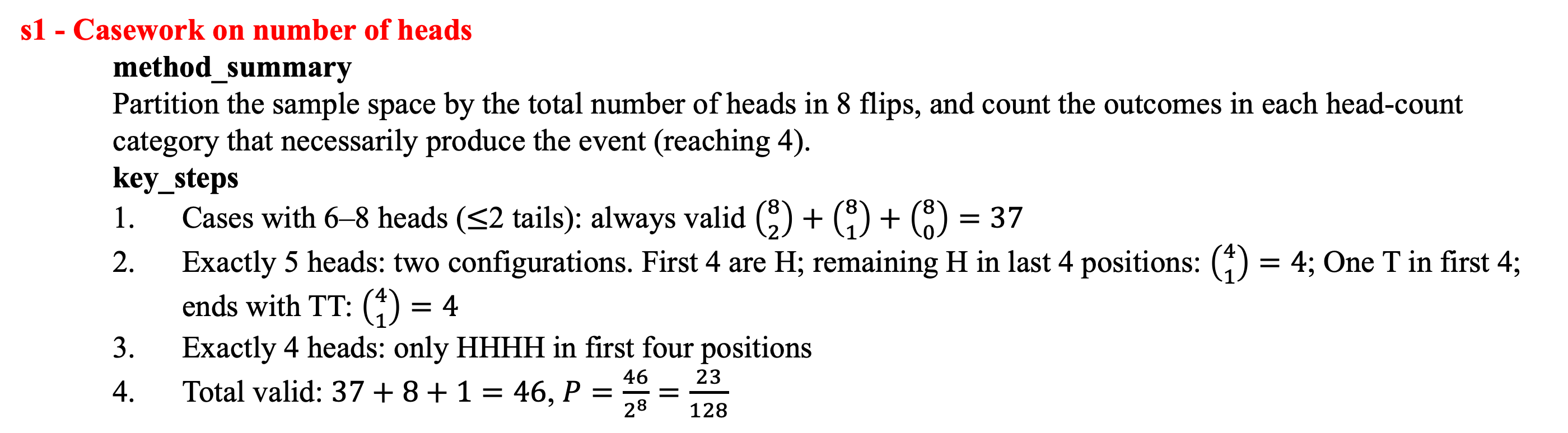}\par\vspace{2pt}
\noindent\includegraphics[width=\linewidth]{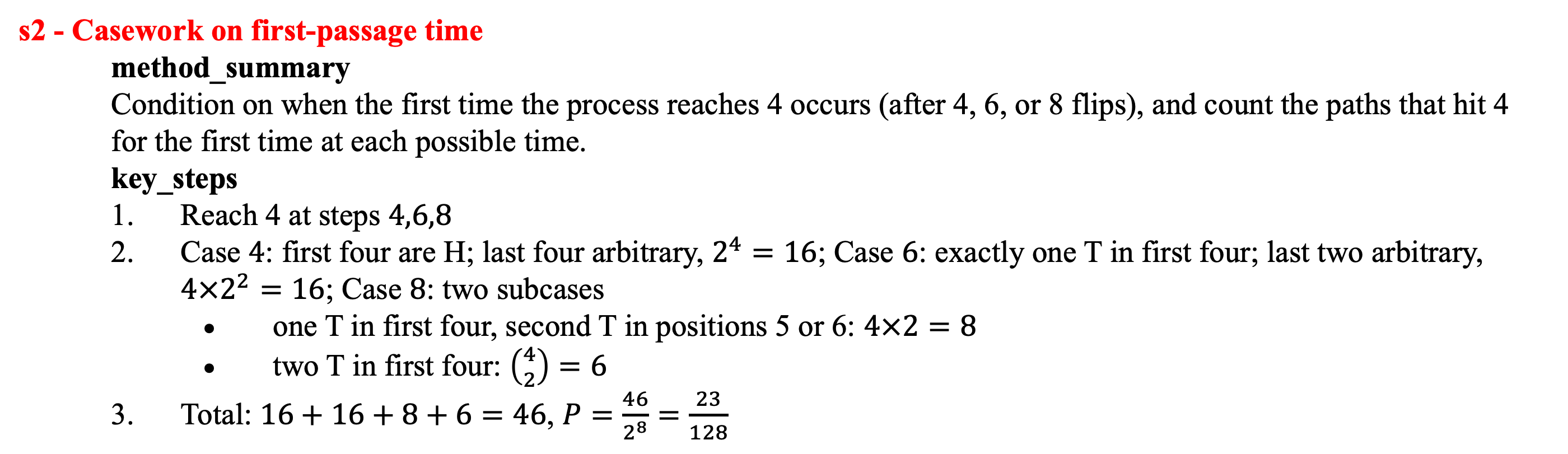}\par\vspace{2pt}
\noindent\includegraphics[width=\linewidth]{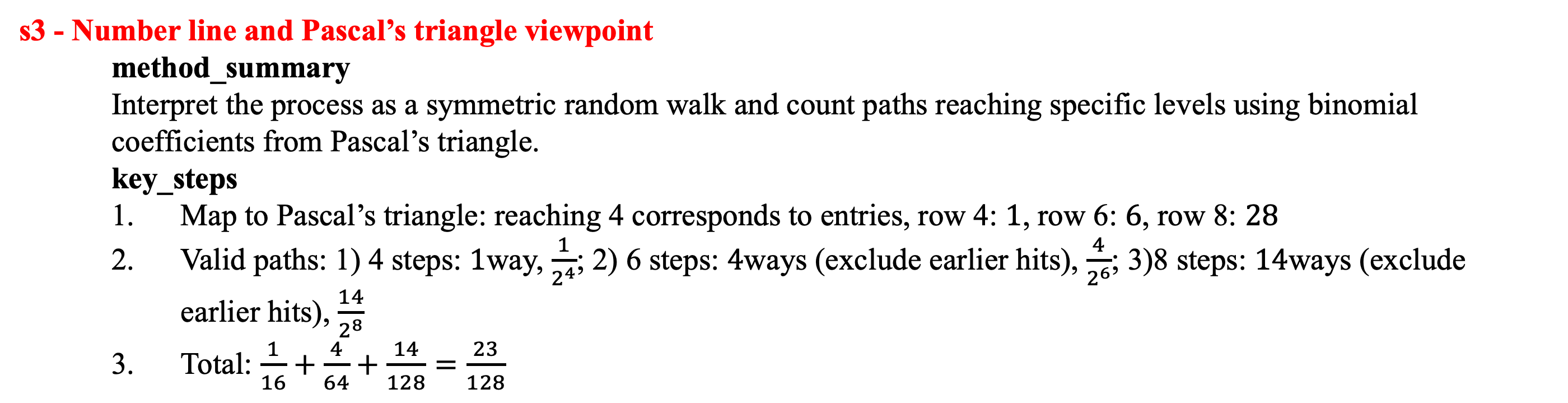}\par\vspace{2pt}
\noindent\includegraphics[width=\linewidth]{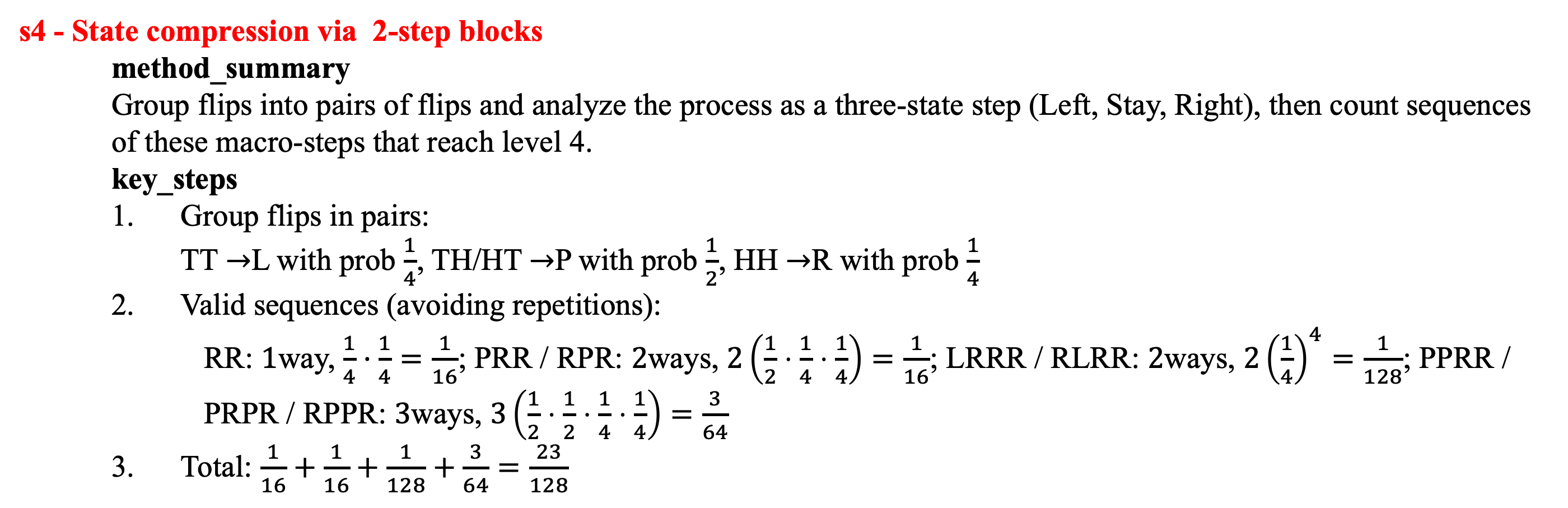}\par\vspace{2pt}
\noindent\includegraphics[width=\linewidth]{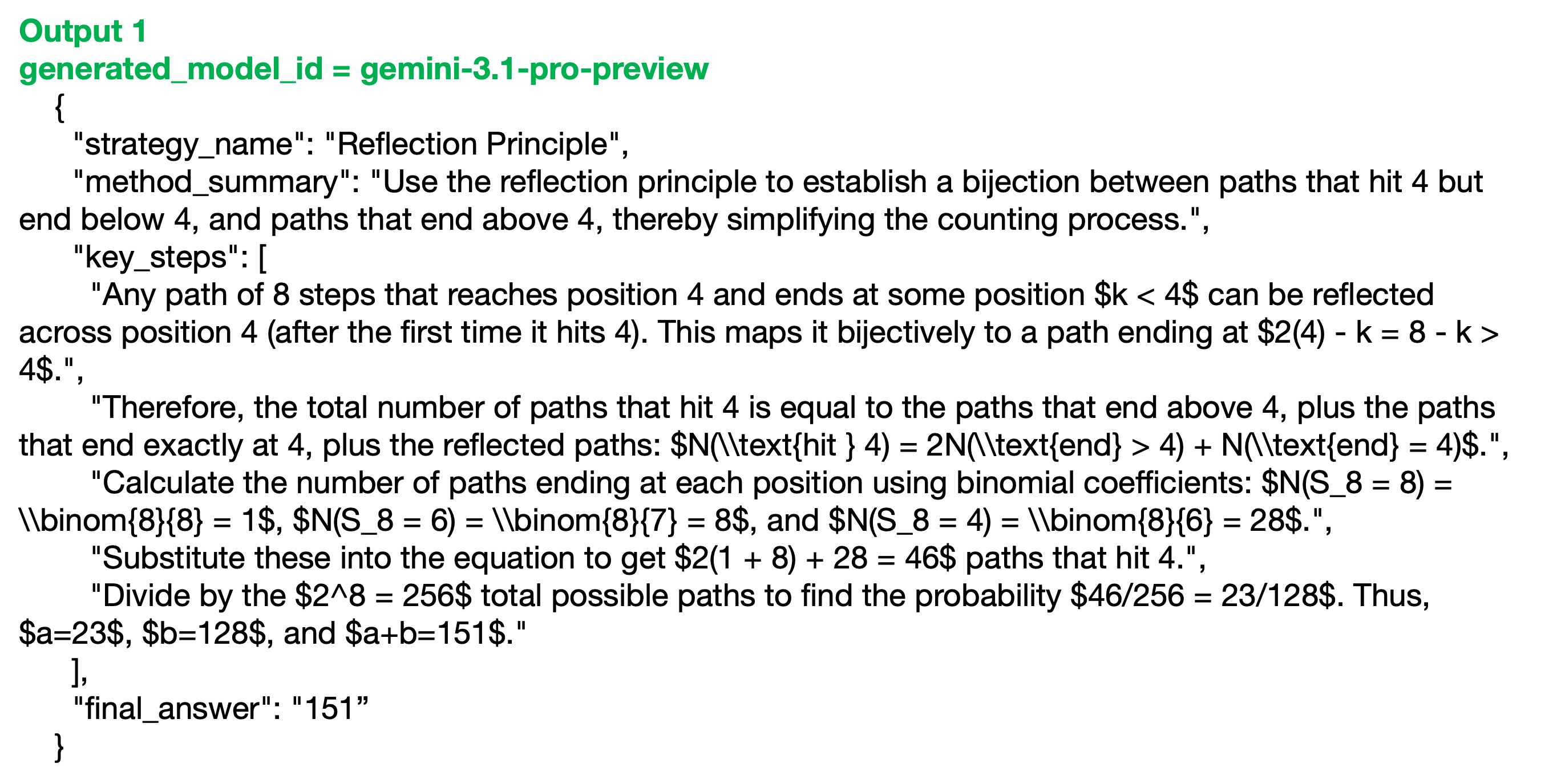}\par\vspace{2pt}
\noindent\includegraphics[width=\linewidth]{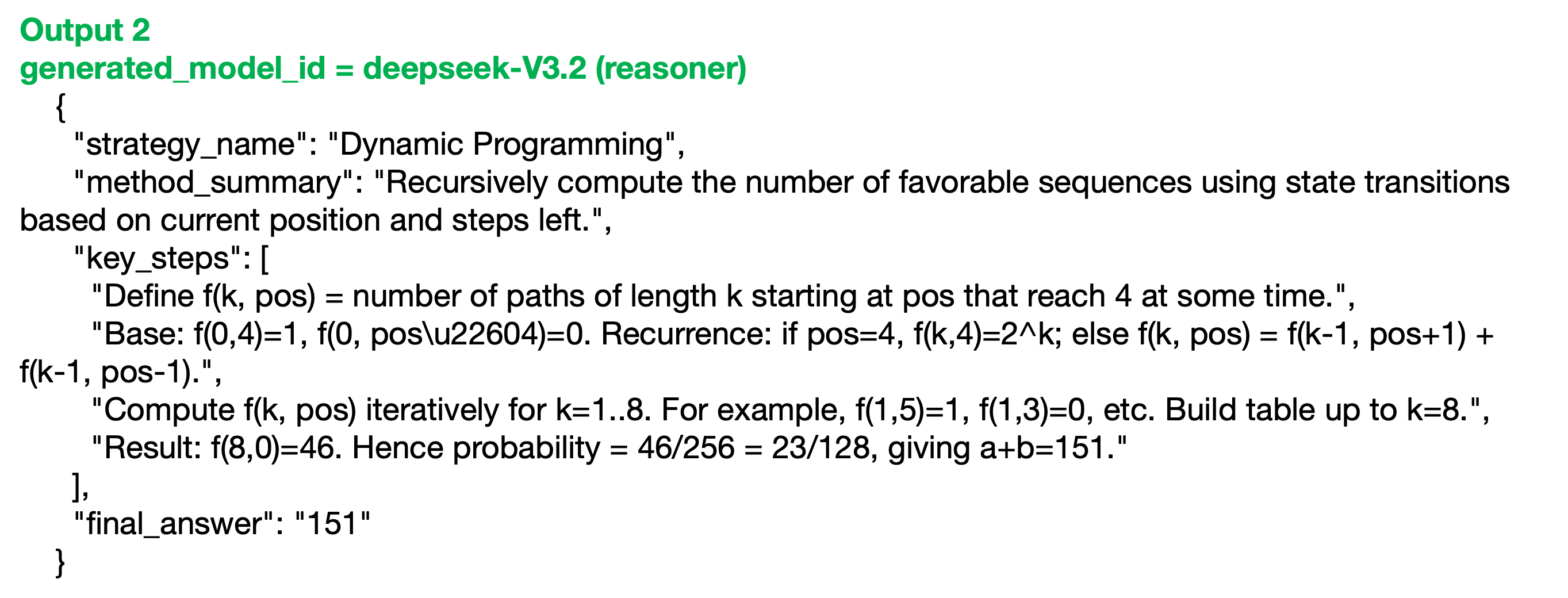}\par\vspace{2pt}
\noindent\includegraphics[width=\linewidth]{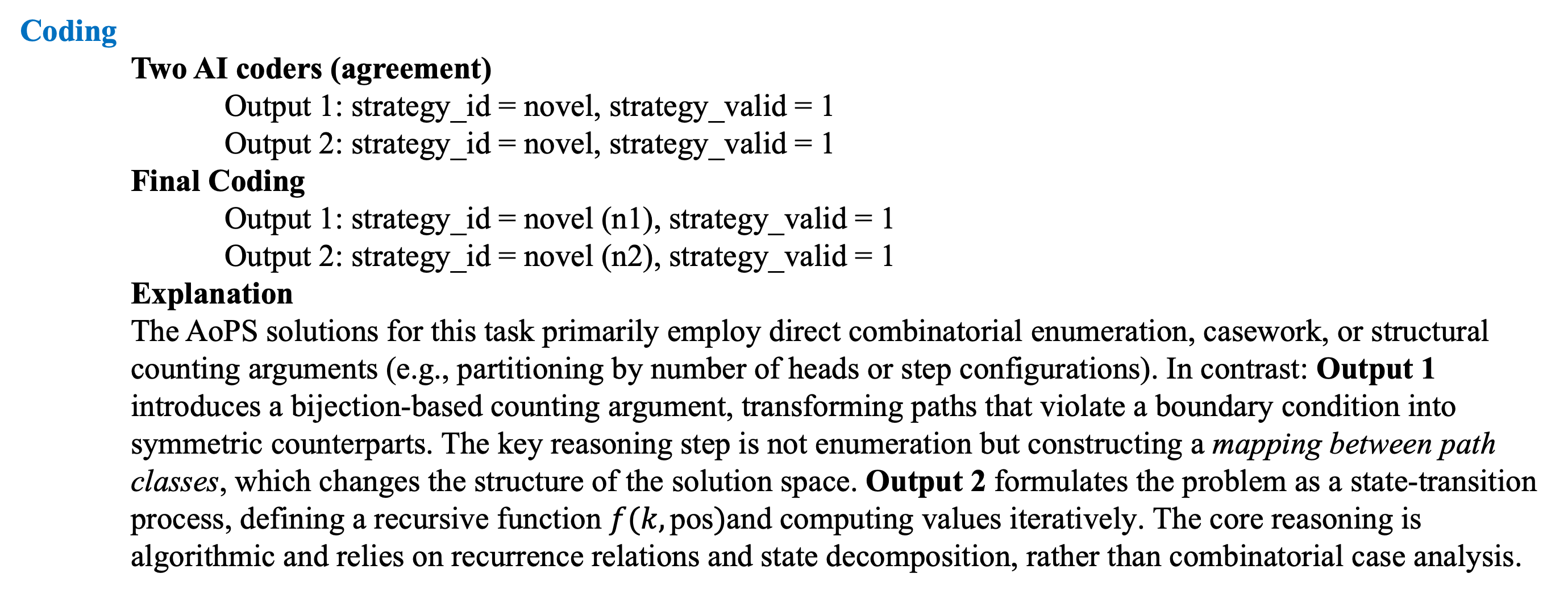}\par\vspace{2pt}

\newpage
\subsection{Example 4: Dominant critical step in multi-stage solutions}
This example illustrates how solutions may involve multiple critical stages, but are assigned a single strategy label based on the dominant reasoning step that determines the solution pathway.

\noindent\includegraphics[width=\linewidth]{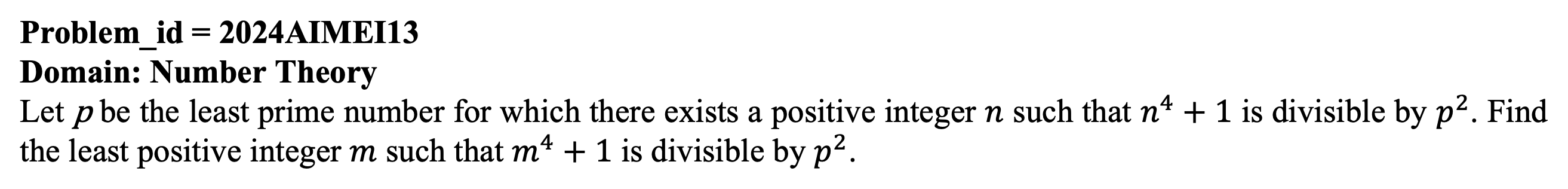}\par\vspace{2pt}
\noindent\includegraphics[width=\linewidth]{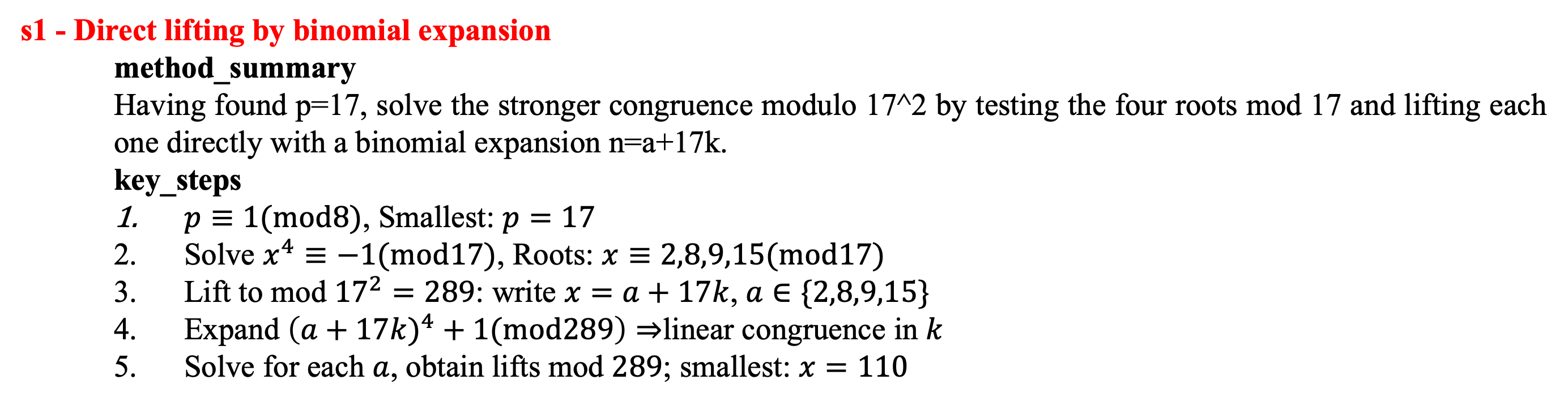}\par\vspace{2pt}
\noindent\includegraphics[width=\linewidth]{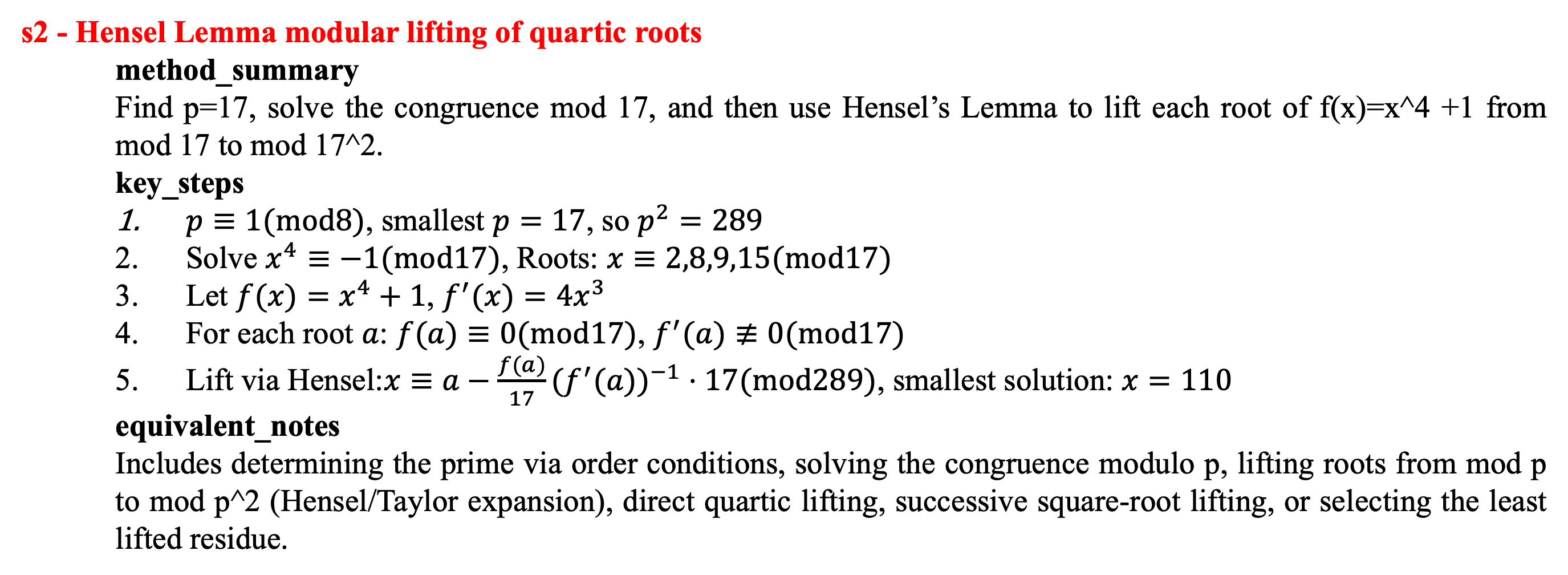}\par\vspace{2pt}
\noindent\includegraphics[width=\linewidth]{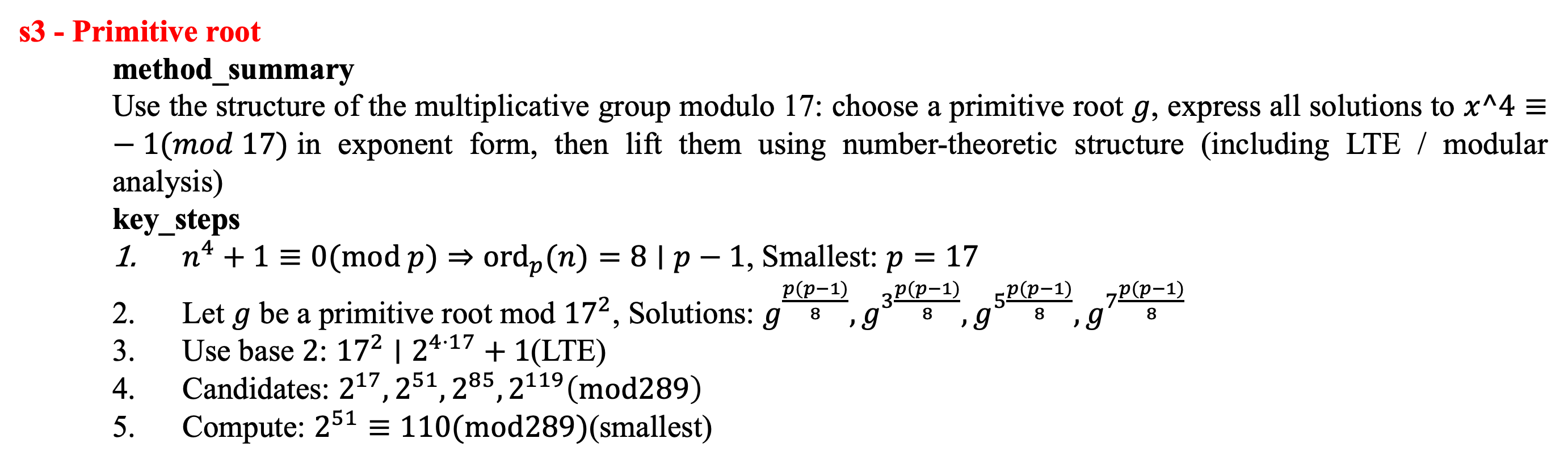}\par\vspace{2pt}
\noindent\includegraphics[width=0.84\linewidth]{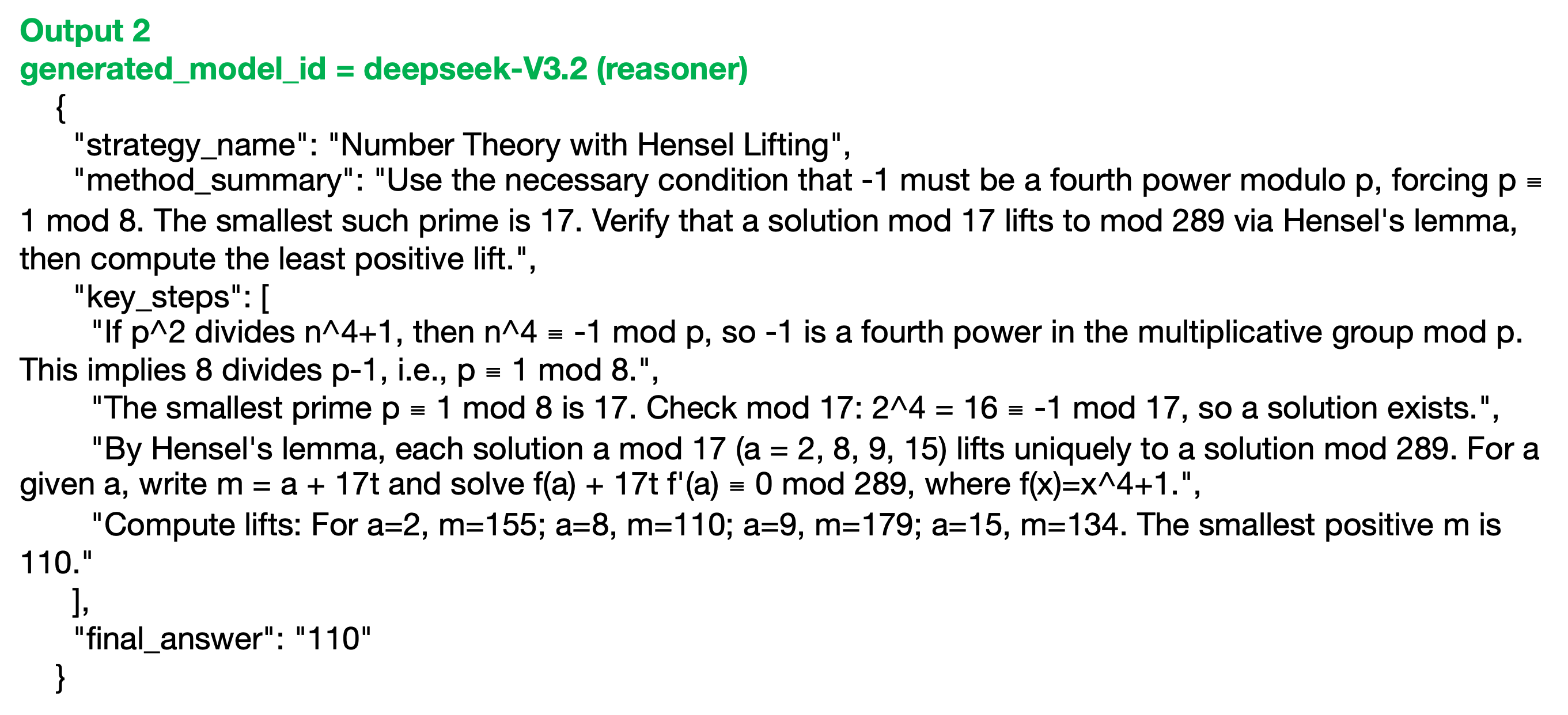}\par\vspace{2pt}
\noindent\includegraphics[width=\linewidth]{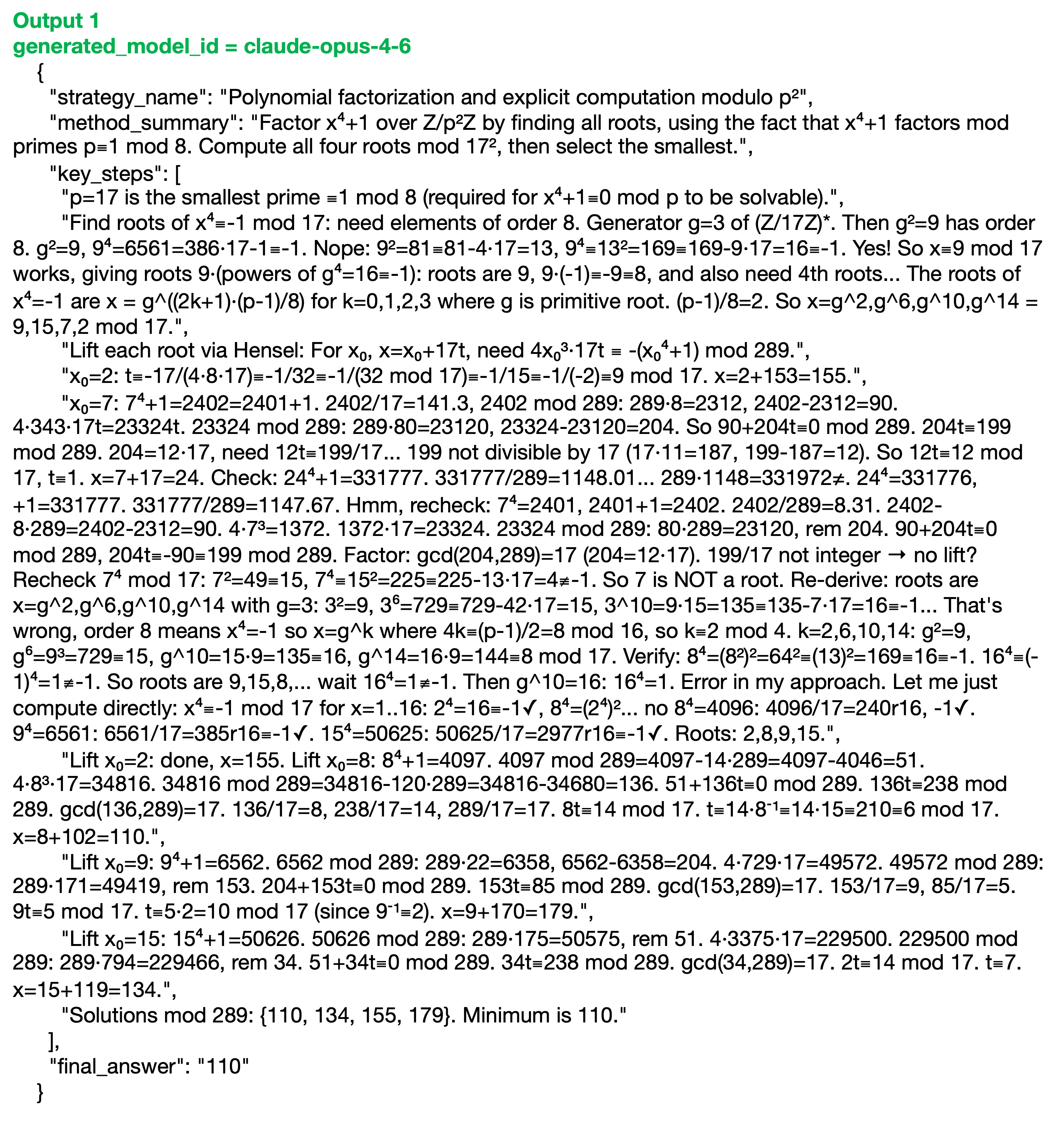}\par\vspace{2pt}
\noindent\includegraphics[width=\linewidth]{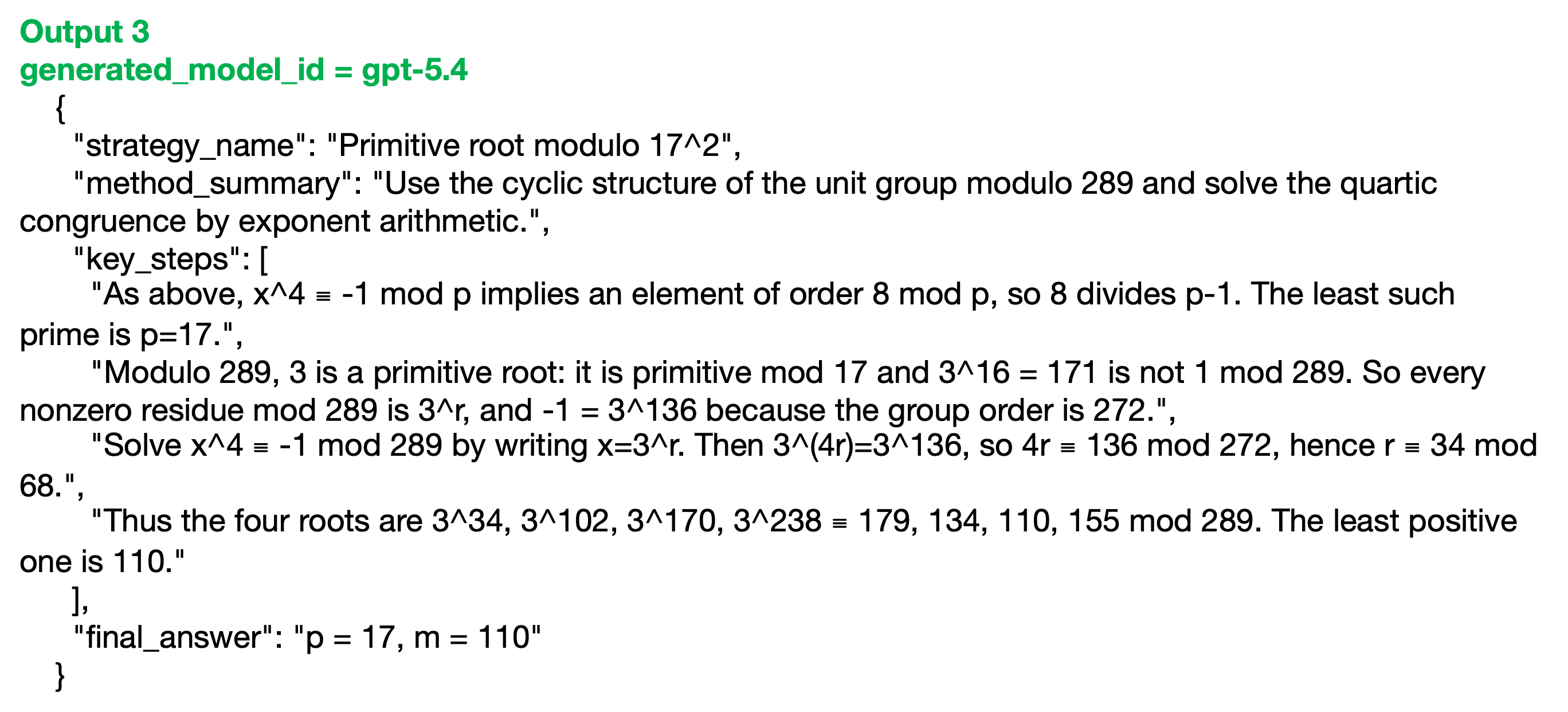}\par\vspace{2pt}
\noindent\includegraphics[width=\linewidth]{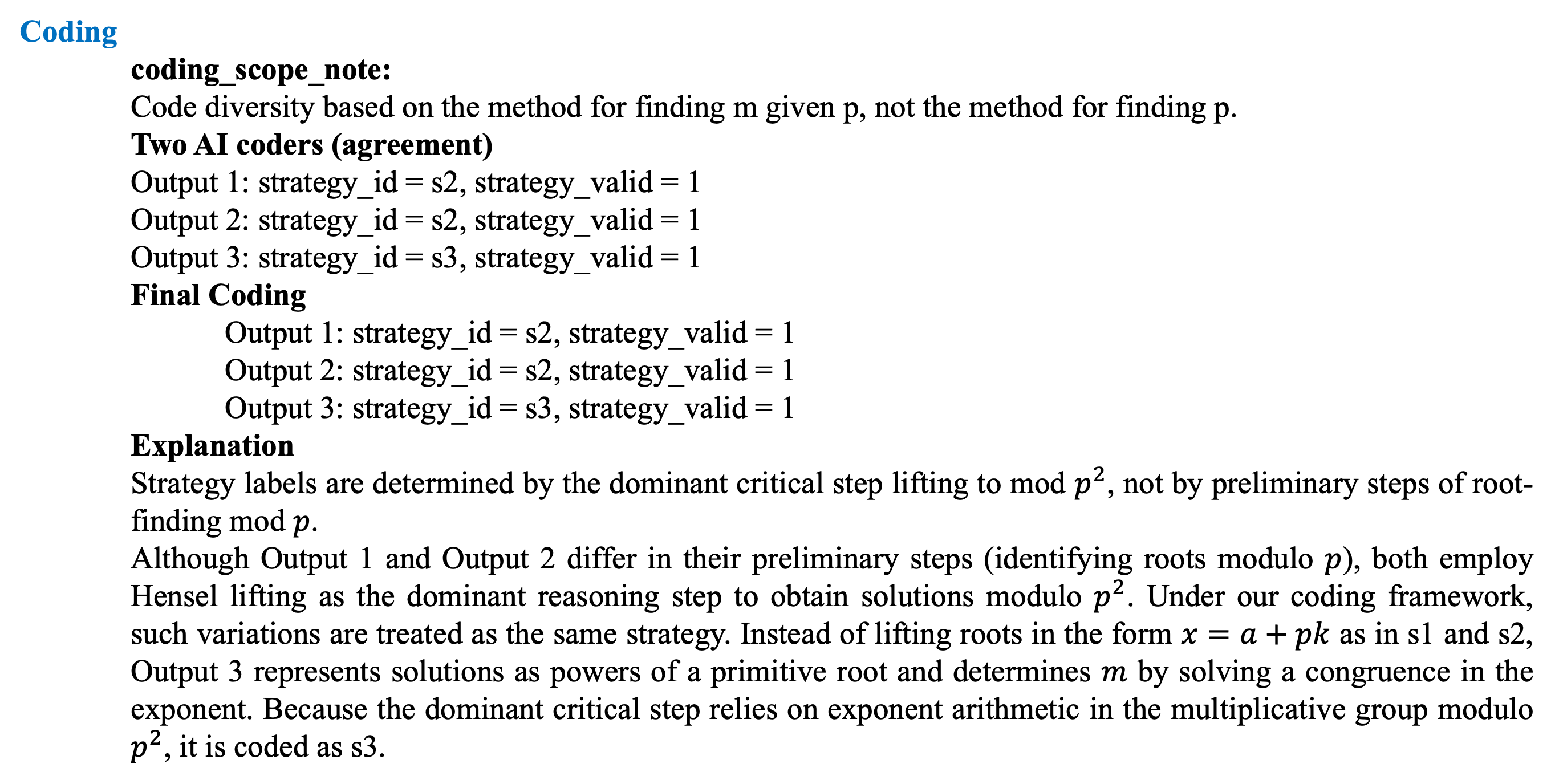}\par\vspace{2pt}

\subsection{Example 5: Human adjudication of AI disagreement}
\label{app:example5}
This example illustrates how human adjudication resolves AI-coder disagreement in boundary cases.

\noindent\includegraphics[width=\linewidth]{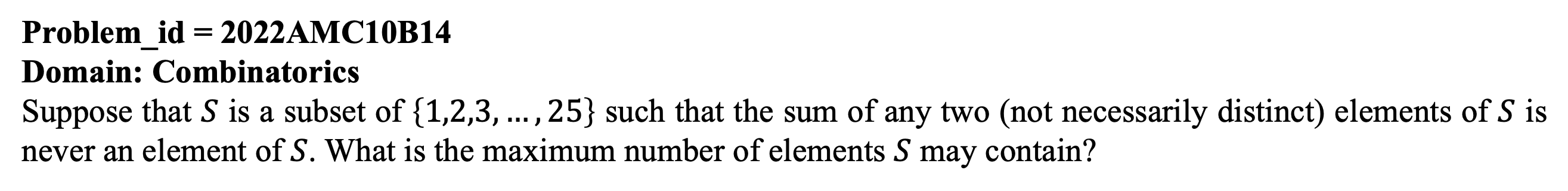}\par\vspace{2pt}
\noindent\includegraphics[width=\linewidth]{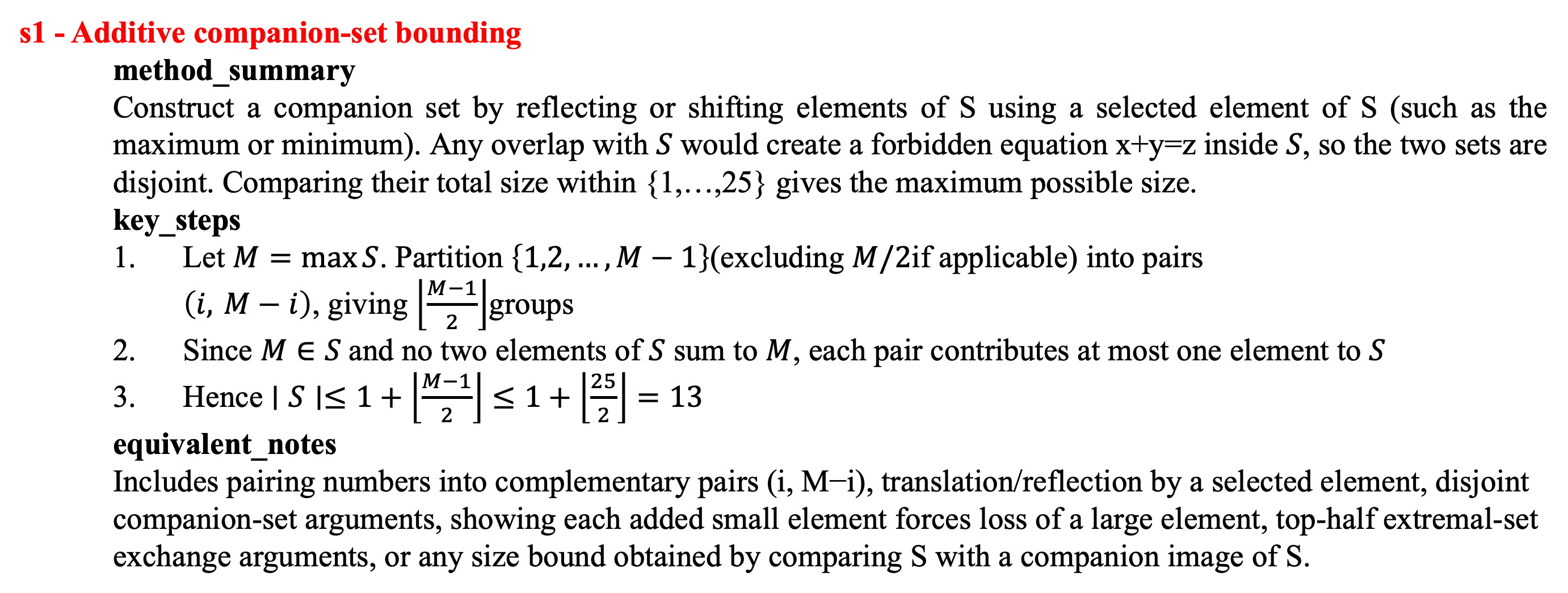}\par\vspace{2pt}
\noindent\includegraphics[width=\linewidth]{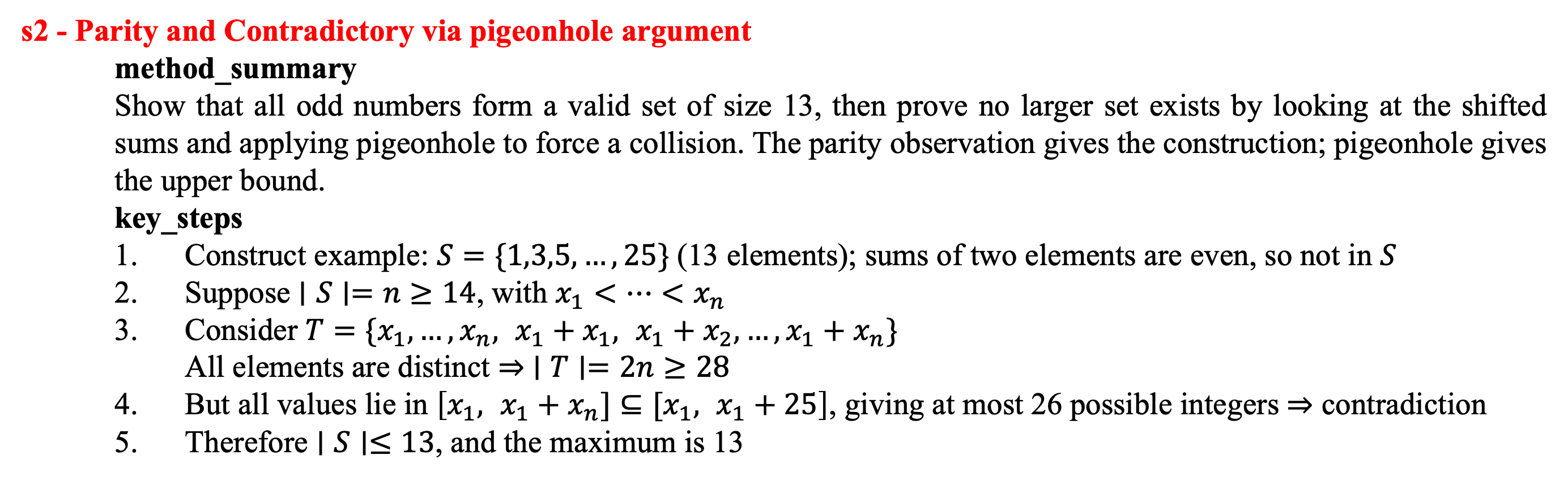}\par\vspace{2pt}
\noindent\includegraphics[width=\linewidth]{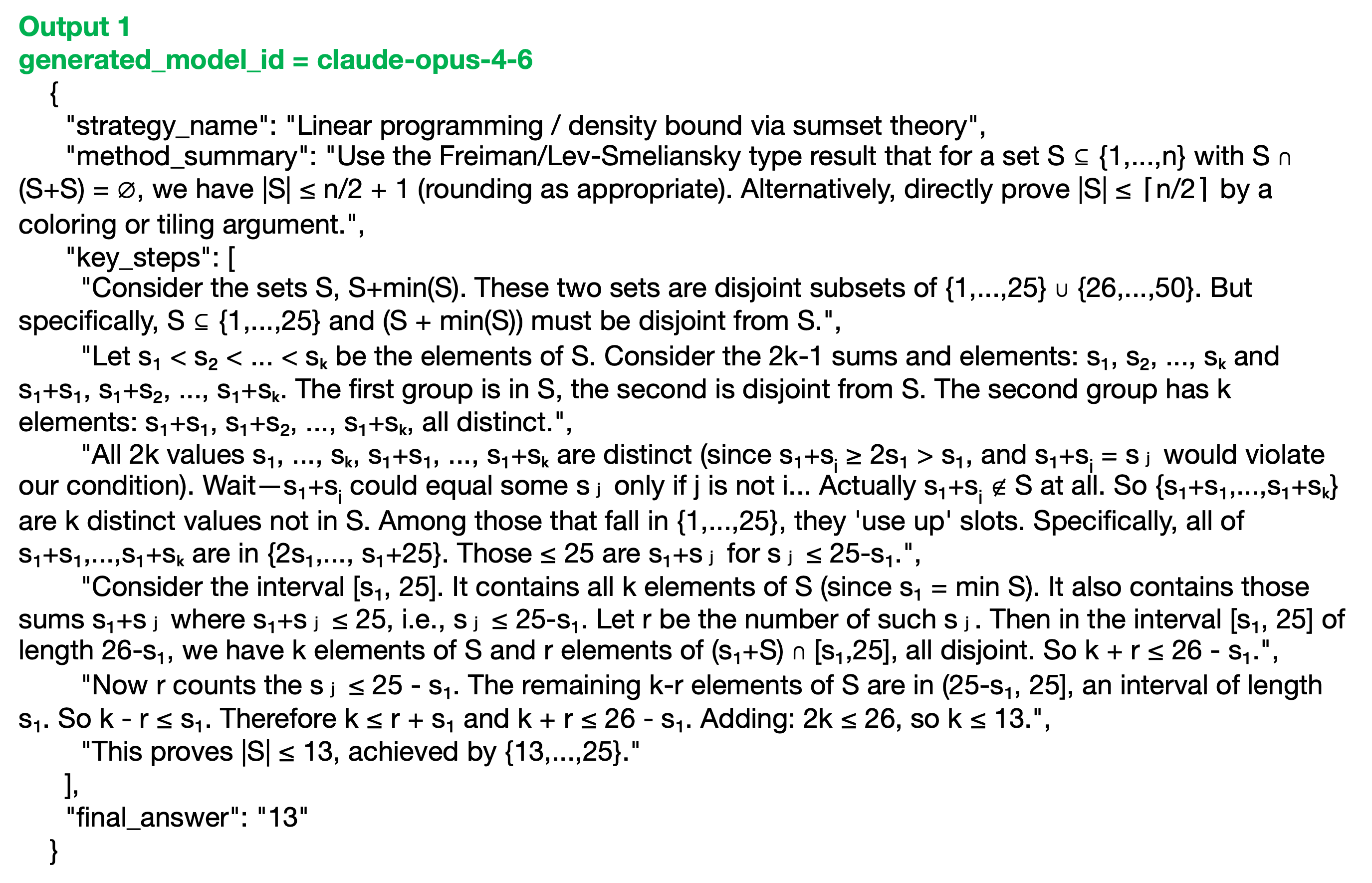}\par\vspace{2pt}
\noindent\includegraphics[width=\linewidth]{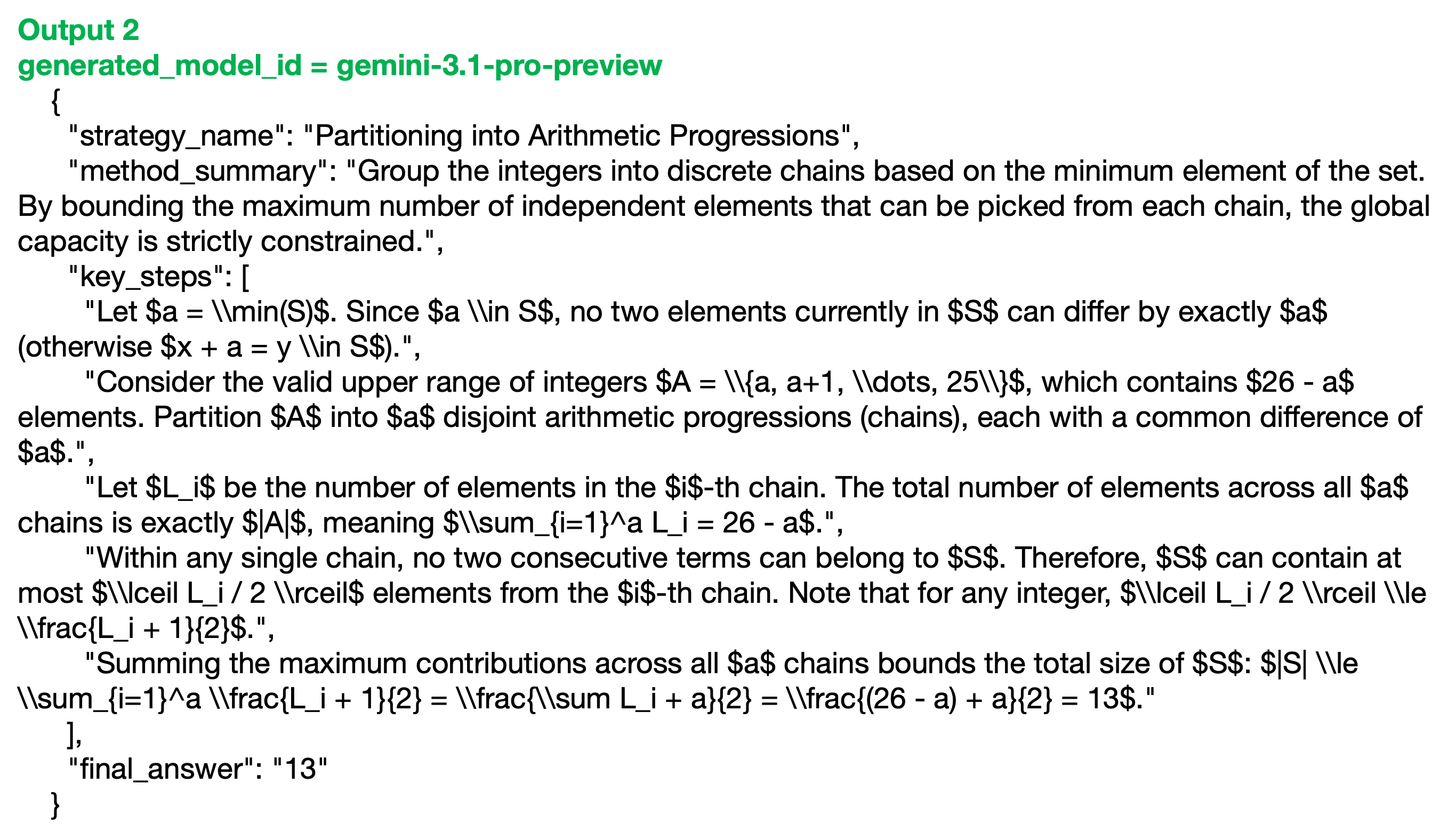}\par\vspace{2pt}
\noindent\includegraphics[width=\linewidth]{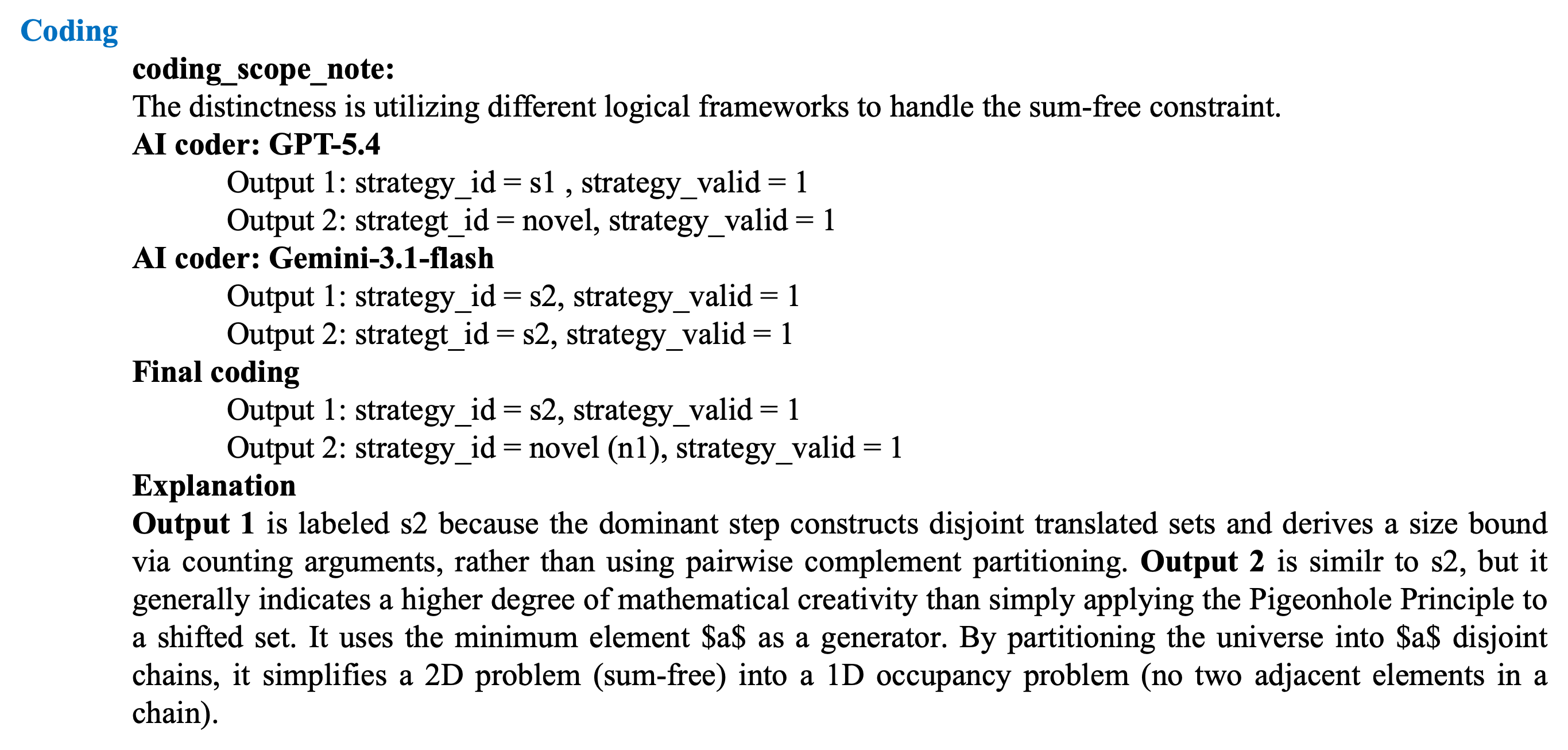}\par\vspace{2pt}

\section{Per-problem strategy inventory}
Table~\ref{tab:appendixinventory} reports, for each of the 80 framework problems, both the finalized strategy inventory and the per-model correctness comparison between \texttt{prompt\_single} and \texttt{prompt\_multi}. The inventory columns give the number of AoPS-reference strategies for that problem, the number of benchmark-novel valid strategies in the finalized inventory, and the resulting total. The model columns use the compact format \texttt{S/M}, where \texttt{S} indicates whether the model solved the problem correctly under \texttt{prompt\_single} and \texttt{M} indicates whether it produced at least one correct strategy under \texttt{prompt\_multi}. This makes it possible to see whether models remain stuck on the same problems or recover additional correct solutions when asked for multiple strategies.

\begin{table}[H]
\centering
\caption{Model-level McNemar summary comparing \texttt{prompt\_single} and \texttt{prompt\_multi} correctness over the 80-problem framework. Counts are paired at the problem level: \texttt{1/0} denotes problems solved only under \texttt{prompt\_single}, and \texttt{0/1} denotes problems solved only under \texttt{prompt\_multi}. We report the two-sided exact McNemar $p$-value.}
\label{tab:appendixmcnemar}
\small
\begin{tabular}{lrrrrrc}
\toprule
Model & 1/1 & 1/0 & 0/1 & 0/0 & Discordant & McNemar $p$ \\
\midrule
GPT & 75 & 2 & 3 & 0 & 5 & 1.0000 \\
Gemini & 80 & 0 & 0 & 0 & 0 & 1.0000 \\
DeepSeek & 77 & 2 & 0 & 1 & 2 & 0.5000 \\
Claude & 68 & 8 & 1 & 3 & 9 & 0.0391 \\
\bottomrule
\end{tabular}
\end{table}

\begingroup
\footnotesize
\setlength{\tabcolsep}{4pt}
\setlength{\LTleft}{0pt}
\setlength{\LTright}{0pt}
\begin{longtable}{llrrrcccc}
\caption{Per-problem finalized strategy inventory for the 80-problem framework, with per-model correctness under \texttt{prompt\_single} and \texttt{prompt\_multi}. ``AoPS'' is the number of reference strategies drawn from the AoPS corpus for that problem, ``Novel'' is the number of benchmark-novel valid strategies in the finalized inventory, and ``Total'' is their sum. In each model column, \texttt{S/M} denotes \texttt{prompt\_single}/\texttt{prompt\_multi} correctness, with \texttt{M=1} if at least one generated strategy for that problem is finally labeled \texttt{result\_correct = 1}.}\label{tab:appendixinventory}\\
\toprule
Problem ID & Domain & AoPS & Novel & Total & GPT S/M & Gemini S/M & DeepSeek S/M & Claude S/M \\
\midrule
\endfirsthead
\toprule
Problem ID & Domain & AoPS & Novel & Total & GPT S/M & Gemini S/M & DeepSeek S/M & Claude S/M \\
\midrule
\endhead
\bottomrule
\endfoot
\input{generated_tables/table_A2_rows.tex}
\end{longtable}
\endgroup

\section{Supplementary AoPS--model gap summary}
The main paper visualizes AoPS--model diversity gaps with paired-gap figures. Table~\ref{tab:appendixgap} provides the corresponding numeric summary for readers who prefer exact values. To avoid redundancy, we keep only the columns that directly support interpretation: analysis group, domain or overall group, model name, number of problems, mean paired gap, bootstrap confidence interval, and paired permutation $p$-value.

\begingroup
\footnotesize
\setlength{\tabcolsep}{4pt}
\setlength{\LTleft}{\fill}
\setlength{\LTright}{\fill}
\begin{longtable}{llcrrrrc}
\caption{Numerical summary of AoPS--model paired diversity gaps.}\label{tab:appendixgap}\\
\toprule
Level & Group & Model & N & Mean Gap & CI$_L$ & CI$_U$ & $p$ \\
\midrule
\endfirsthead
\toprule
Level & Group & Model & N & Mean Gap & CI$_L$ & CI$_U$ & $p$ \\
\midrule
\endhead
\bottomrule
\endfoot
\input{generated_tables/table_A3_rows.tex}
\end{longtable}
\endgroup

\newpage
\section{Additional novelty overlap figures}
The main text reports the domain distribution of AoPS-novel strategies and the overall repeated-run saturation trend. The figures in this section provide a finer-grained view of novelty overlap.

\begin{figure}[H]
    \centering
    \includegraphics[width=0.65\linewidth]{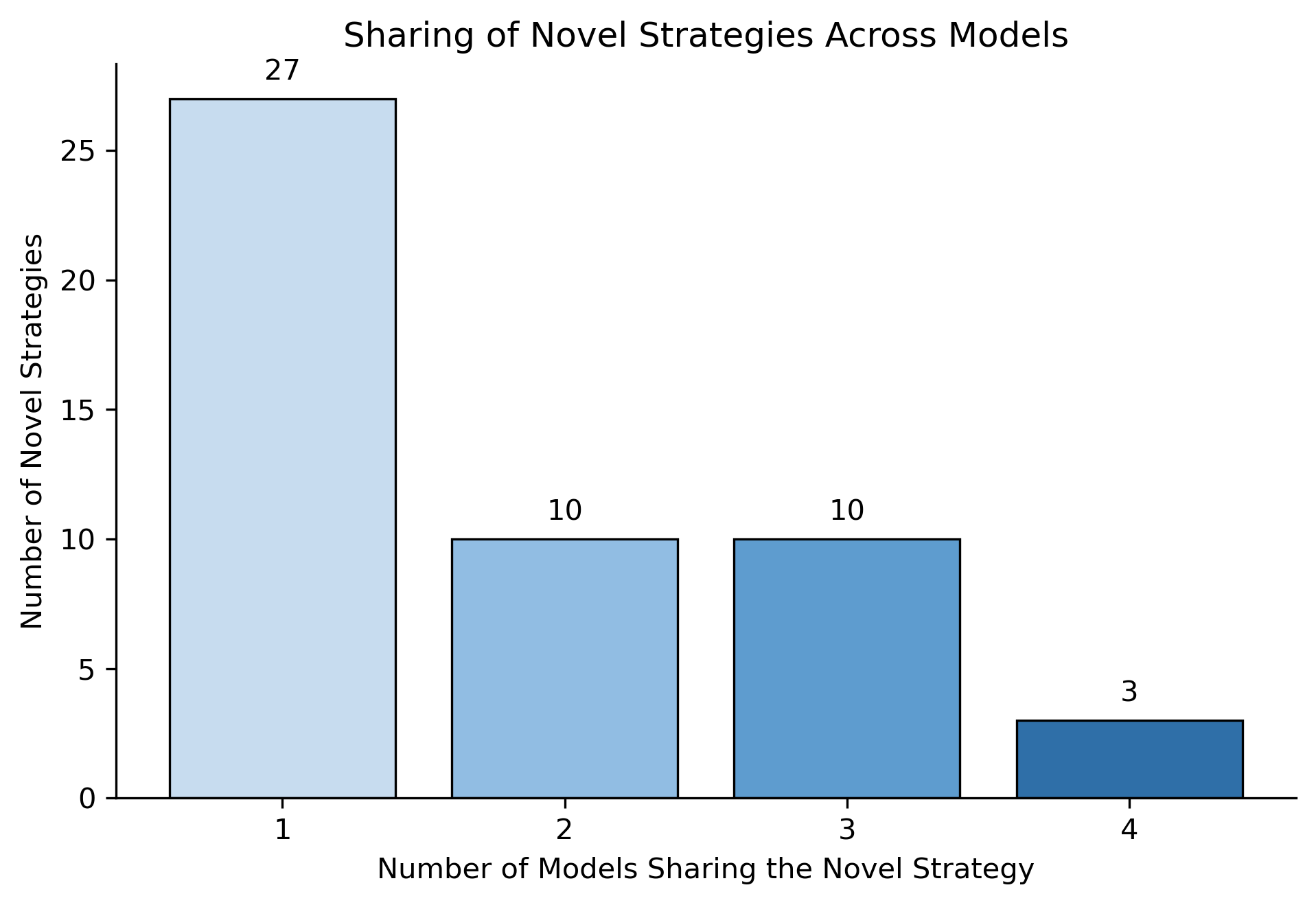}
    \caption{Sharing patterns of novel strategies across models.}
    \vspace{2pt}
    \includegraphics[width=\linewidth]{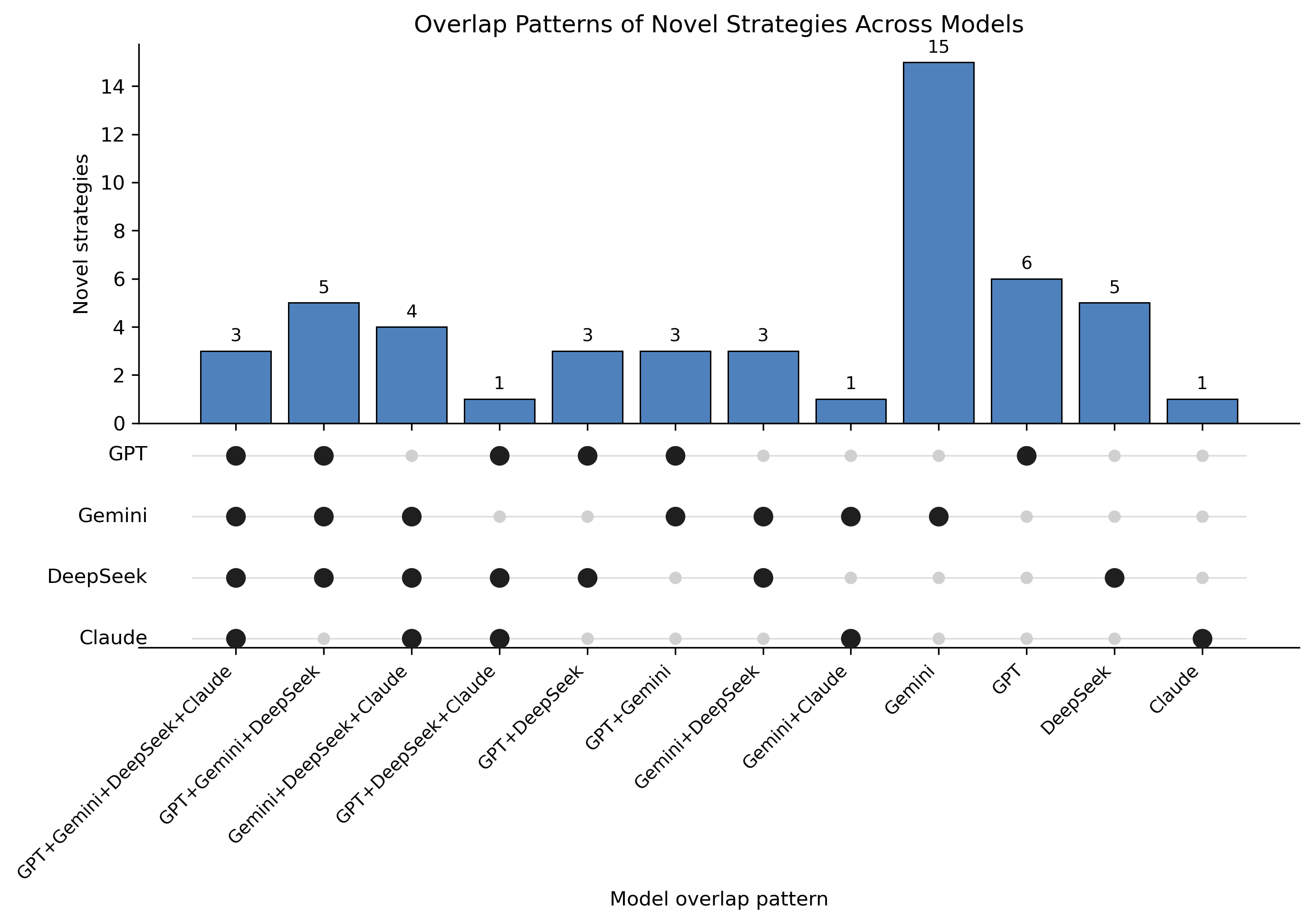}
    \caption{Overlap structure of novel strategies across models. Each column corresponds to one model-overlap pattern, and the bar above it gives the number of AoPS-novel strategies with that exact overlap structure.}
    \label{fig:upset}
\end{figure}

\clearpage
\section{Repeated-run robustness details}
The repeated-run analysis on the 20-problem subset is intended as a robustness and saturation check rather than a second framework instantiation. We therefore report the quantities most directly relevant to interpretation: the number of valid distinct strategies found in each run, the size of their union, and the number of newly discovered strategies added by each additional run.

\begin{table}[H]
\centering
\caption{Three-run comparison on the 20-problem subset. The AoPS reference pool for this subset contains 55 strategies. ``Full'' is the original full-framework slice restricted to the same 20 problems; ``U(F+R1)'' is the union of Full and run 1; ``U(All)'' is the union across Full, run 1, and run 2. For each stage, we report the total number of distinct strategies (Tot), the number of AoPS-reference strategies recovered by the model (AoPS Rec.), and AoPS coverage relative to the fixed 55-strategy reference pool.}
\label{tab:runcompare}
\footnotesize
\setlength{\tabcolsep}{2.5pt}
\resizebox{0.84\columnwidth}{!}{%
\begin{tabular}{lrrrrrrrrr}
\toprule
& \multicolumn{3}{c}{Full} & \multicolumn{3}{c}{U(F+R1)} & \multicolumn{3}{c}{U(All)} \\
\cmidrule(lr){2-4}\cmidrule(lr){5-7}\cmidrule(lr){8-10}
Model & Tot & AoPS Rec. & Cov. & Tot & AoPS Rec. & Cov. & Tot & AoPS Rec. & Cov. \\
\midrule
Claude & 28 & 25 & 45.5\% & 30 & 27 & 49.1\% & 33 & 30 & 54.5\% \\
DeepSeek & 35 & 28 & 50.9\% & 43 & 36 & 65.5\% & 46 & 38 & 69.1\% \\
GPT & 35 & 28 & 50.9\% & 42 & 34 & 61.8\% & 44 & 35 & 63.6\% \\
Gemini & 46 & 35 & 63.6\% & 52 & 37 & 67.3\% & 54 & 39 & 70.9\% \\
\bottomrule
\end{tabular}
\par}
\end{table}

\begin{table*}[htbp]
\centering
\caption{Model-level paired gains under repeated sampling on the 20-problem subset. For each model, statistics are computed over the same 20 problems. ``AoPS Rec.'' reports the number of AoPS-reference strategies recovered at each stage, while ``Total'' reports the total number of distinct strategies. Mean gains, confidence intervals, and one-sided paired permutation $p$-values are computed at the paired problem level within model.}
\label{tab:repeatgainmodel}
\scriptsize
\setlength{\tabcolsep}{3pt}
\resizebox{\textwidth}{!}{%
\begin{tabular}{llcccccccc}
\toprule
& & \multicolumn{4}{c}{AoPS Rec.} & \multicolumn{4}{c}{Total Distinct Strategies} \\
\cmidrule(lr){3-6}\cmidrule(lr){7-10}
Model & Comparison & Count & Gain & 95\% CI & Perm. $p$ & Count & Gain & 95\% CI & Perm. $p$ \\
\midrule
Claude & Full $\rightarrow$ Full+Run1 & 25 $\rightarrow$ 27 & +0.10 & [0.00, 0.25] & 0.251 & 28 $\rightarrow$ 30 & +0.10 & [0.00, 0.25] & 0.251 \\
 & Full+Run1 $\rightarrow$ All & 27 $\rightarrow$ 30 & +0.15 & [0.00, 0.30] & 0.124 & 30 $\rightarrow$ 33 & +0.15 & [0.00, 0.30] & 0.124 \\
 & Full $\rightarrow$ All & 25 $\rightarrow$ 30 & +0.25 & [0.10, 0.45] & 0.031 & 28 $\rightarrow$ 33 & +0.25 & [0.10, 0.45] & 0.031 \\
\midrule
DeepSeek & Full $\rightarrow$ Full+Run1 & 28 $\rightarrow$ 36 & +0.40 & [0.15, 0.65] & 0.008 & 35 $\rightarrow$ 43 & +0.40 & [0.15, 0.65] & 0.008 \\
 & Full+Run1 $\rightarrow$ All & 36 $\rightarrow$ 38 & +0.10 & [0.00, 0.25] & 0.248 & 43 $\rightarrow$ 46 & +0.15 & [0.00, 0.30] & 0.123 \\
 & Full $\rightarrow$ All & 28 $\rightarrow$ 38 & +0.50 & [0.20, 0.80] & 0.004 & 35 $\rightarrow$ 46 & +0.55 & [0.25, 0.85] & 0.002 \\
\midrule
GPT & Full $\rightarrow$ Full+Run1 & 28 $\rightarrow$ 34 & +0.30 & [0.10, 0.55] & 0.031 & 35 $\rightarrow$ 42 & +0.35 & [0.10, 0.60] & 0.016 \\
 & Full+Run1 $\rightarrow$ All & 34 $\rightarrow$ 35 & +0.05 & [0.00, 0.15] & 0.500 & 42 $\rightarrow$ 44 & +0.10 & [0.00, 0.25] & 0.250 \\
 & Full $\rightarrow$ All & 28 $\rightarrow$ 35 & +0.35 & [0.10, 0.60] & 0.016 & 35 $\rightarrow$ 44 & +0.45 & [0.20, 0.75] & 0.008 \\
\midrule
Gemini & Full $\rightarrow$ Full+Run1 & 35 $\rightarrow$ 37 & +0.10 & [0.00, 0.25] & 0.249 & 46 $\rightarrow$ 52 & +0.30 & [0.10, 0.55] & 0.031 \\
 & Full+Run1 $\rightarrow$ All & 37 $\rightarrow$ 39 & +0.10 & [0.00, 0.25] & 0.250 & 52 $\rightarrow$ 54 & +0.10 & [0.00, 0.25] & 0.250 \\
 & Full $\rightarrow$ All & 35 $\rightarrow$ 39 & +0.20 & [0.00, 0.45] & 0.125 & 46 $\rightarrow$ 54 & +0.40 & [0.15, 0.70] & 0.015 \\
\bottomrule
\end{tabular}
\par}
\end{table*}

\end{document}

%% file: generated_tables/table_A2_rows.tex
2017AIMEI5 & Algebra & 3 & 0 & 3 & 1/1 & 1/1 & 1/1 & 1/1 \\
2017AMC12A15 & Algebra & 1 & 2 & 3 & 1/1 & 1/1 & 1/1 & 1/1 \\
2017AMC12A23 & Algebra & 2 & 0 & 2 & 1/1 & 1/1 & 1/1 & 1/1 \\
2018AMC10A12 & Algebra & 4 & 0 & 4 & 1/1 & 1/1 & 1/1 & 1/1 \\
2018AMC12A14 & Algebra & 2 & 0 & 2 & 1/1 & 1/1 & 1/1 & 1/1 \\
2019AMC10A15 & Algebra & 2 & 0 & 2 & 1/1 & 1/1 & 1/1 & 1/1 \\
2019AMC12A21 & Algebra & 2 & 0 & 2 & 1/1 & 1/1 & 1/1 & 1/1 \\
2020AMC10B24 & Algebra & 3 & 0 & 3 & 1/1 & 1/1 & 1/1 & 1/1 \\
2021AMC12B21 & Algebra & 4 & 1 & 5 & 1/1 & 1/1 & 1/1 & 1/1 \\
2024AMC10A13 & Algebra & 2 & 0 & 2 & 1/1 & 1/1 & 1/1 & 1/1 \\
2024AMC10B23 & Algebra & 3 & 0 & 3 & 1/1 & 1/1 & 1/1 & 1/1 \\
2024AMC12A13 & Algebra & 4 & 0 & 4 & 1/1 & 1/1 & 1/1 & 1/1 \\
2025AMC10A5 & Algebra & 4 & 0 & 4 & 1/1 & 1/1 & 1/1 & 1/1 \\
2025AMC12A25 & Algebra & 1 & 0 & 1 & 1/1 & 1/1 & 1/0 & 0/0 \\
2016AIMEII13 & Combinatorics & 2 & 1 & 3 & 1/0 & 1/1 & 1/0 & 1/1 \\
2016AMC10A18 & Combinatorics & 3 & 1 & 4 & 1/0 & 1/1 & 1/1 & 1/1 \\
2017AIMEII11 & Combinatorics & 1 & 4 & 5 & 1/1 & 1/1 & 1/1 & 1/1 \\
2017AIMEII14 & Combinatorics & 3 & 0 & 3 & 1/1 & 1/1 & 1/1 & 1/1 \\
2017AMC10B13 & Combinatorics & 3 & 0 & 3 & 1/1 & 1/1 & 1/1 & 1/1 \\
2018AMC10A11 & Combinatorics & 3 & 0 & 3 & 1/1 & 1/1 & 1/1 & 1/1 \\
2019AMC10A17 & Combinatorics & 2 & 0 & 2 & 1/1 & 1/1 & 1/1 & 1/1 \\
2020AMC10B23 & Combinatorics & 3 & 1 & 4 & 1/1 & 1/1 & 1/1 & 1/1 \\
2022AMC10A14 & Combinatorics & 1 & 2 & 3 & 1/1 & 1/1 & 1/1 & 1/0 \\
2022AMC10A24 & Combinatorics & 5 & 1 & 6 & 1/1 & 1/1 & 1/1 & 1/0 \\
2022AMC10B14 & Combinatorics & 2 & 1 & 3 & 1/1 & 1/1 & 1/1 & 1/1 \\
2022AMC10B18 & Combinatorics & 3 & 1 & 4 & 1/1 & 1/1 & 1/1 & 1/1 \\
2023AMC10A18 & Combinatorics & 3 & 1 & 4 & 1/1 & 1/1 & 1/1 & 1/1 \\
2023AMC12A24 & Combinatorics & 3 & 2 & 5 & 1/1 & 1/1 & 1/1 & 1/1 \\
2023AMC12B5 & Combinatorics & 2 & 2 & 4 & 1/1 & 1/1 & 1/1 & 1/1 \\
2024AIMEI11 & Combinatorics & 1 & 2 & 3 & 0/1 & 1/1 & 1/1 & 0/1 \\
2024AMC10B12 & Combinatorics & 1 & 0 & 1 & 1/1 & 1/1 & 1/1 & 1/1 \\
2025AIMEI13 & Combinatorics & 1 & 0 & 1 & 1/1 & 1/1 & 1/1 & 0/0 \\
2025AMC12B17 & Combinatorics & 1 & 1 & 2 & 0/1 & 1/1 & 0/0 & 0/0 \\
2016AMC10B23 & Geometry & 3 & 2 & 5 & 1/1 & 1/1 & 1/1 & 1/1 \\
2017AMC10B22 & Geometry & 2 & 0 & 2 & 1/1 & 1/1 & 1/1 & 1/1 \\
2018AMC10B24 & Geometry & 3 & 1 & 4 & 1/1 & 1/1 & 1/1 & 1/0 \\
2018AMC12B21 & Geometry & 2 & 1 & 3 & 1/1 & 1/1 & 1/1 & 1/1 \\
2019AMC10B16 & Geometry & 2 & 3 & 5 & 1/1 & 1/1 & 1/1 & 1/1 \\
2019AMC12B25 & Geometry & 3 & 0 & 3 & 0/1 & 1/1 & 1/1 & 1/1 \\
2020AMC12A12 & Geometry & 5 & 0 & 5 & 1/1 & 1/1 & 1/1 & 1/1 \\
2020AMC12B12 & Geometry & 6 & 0 & 6 & 1/1 & 1/1 & 1/1 & 1/1 \\
2021AMC12B22 & Geometry & 4 & 0 & 4 & 1/1 & 1/1 & 1/1 & 1/1 \\
2021AMC12B24 & Geometry & 5 & 1 & 6 & 1/1 & 1/1 & 1/1 & 1/1 \\
2022AMC10B20 & Geometry & 3 & 1 & 4 & 1/1 & 1/1 & 1/1 & 1/1 \\
2022AMC10B22 & Geometry & 1 & 0 & 1 & 1/1 & 1/1 & 1/1 & 1/0 \\
2023AMC12B25 & Geometry & 2 & 0 & 2 & 1/1 & 1/1 & 1/1 & 1/1 \\
2025AIMEII14 & Geometry & 3 & 0 & 3 & 1/1 & 1/1 & 1/1 & 1/0 \\
2025AMC12B10 & Geometry & 6 & 0 & 6 & 1/1 & 1/1 & 1/1 & 1/0 \\
2025AMC12B25 & Geometry & 3 & 0 & 3 & 1/1 & 1/1 & 1/1 & 1/1 \\
2016AMC10A25 & Number Theory & 2 & 0 & 2 & 1/1 & 1/1 & 1/1 & 1/1 \\
2017AIMEI12 & Number Theory & 4 & 1 & 5 & 1/1 & 1/1 & 1/1 & 1/1 \\
2017AMC10B14 & Number Theory & 2 & 0 & 2 & 1/1 & 1/1 & 1/1 & 1/1 \\
2017AMC10B25 & Number Theory & 4 & 0 & 4 & 1/1 & 1/1 & 1/1 & 1/1 \\
2018AIMEI12 & Number Theory & 5 & 0 & 5 & 1/1 & 1/1 & 1/1 & 1/1 \\
2018AIMEI6 & Number Theory & 2 & 0 & 2 & 1/1 & 1/1 & 1/1 & 1/1 \\
2018AMC10A18 & Number Theory & 4 & 0 & 4 & 1/1 & 1/1 & 1/1 & 1/1 \\
2019AMC10A18 & Number Theory & 2 & 0 & 2 & 1/1 & 1/1 & 1/1 & 1/1 \\
2019AMC10A25 & Number Theory & 2 & 0 & 2 & 1/1 & 1/1 & 1/1 & 1/1 \\
2020AMC10A22 & Number Theory & 2 & 0 & 2 & 1/1 & 1/1 & 1/1 & 1/1 \\
2020AMC10A24 & Number Theory & 3 & 0 & 3 & 1/1 & 1/1 & 1/1 & 1/1 \\
2020AMC10B22 & Number Theory & 4 & 0 & 4 & 1/1 & 1/1 & 1/1 & 1/0 \\
2021AMC12A25 & Number Theory & 3 & 2 & 5 & 1/1 & 1/1 & 1/1 & 1/0 \\
2023AMC12A20 & Number Theory & 3 & 0 & 3 & 1/1 & 1/1 & 1/1 & 1/1 \\
2024AIMEI13 & Number Theory & 3 & 1 & 4 & 1/1 & 1/1 & 1/1 & 1/1 \\
2016AMC10A17 & Probability & 3 & 0 & 3 & 1/1 & 1/1 & 1/1 & 1/1 \\
2016AMC12A19 & Probability & 4 & 3 & 7 & 1/1 & 1/1 & 1/1 & 1/1 \\
2016AMC12A23 & Probability & 3 & 2 & 5 & 1/1 & 1/1 & 1/1 & 1/1 \\
2017AMC12A10 & Probability & 2 & 2 & 4 & 1/1 & 1/1 & 1/1 & 1/1 \\
2017AMC12A22 & Probability & 1 & 2 & 3 & 1/1 & 1/1 & 1/1 & 1/1 \\
2018AMC12A24 & Probability & 2 & 0 & 2 & 1/1 & 1/1 & 1/1 & 1/1 \\
2020AMC10A13 & Probability & 2 & 1 & 3 & 1/1 & 1/1 & 1/1 & 1/1 \\
2020AMC10A25 & Probability & 1 & 1 & 2 & 1/1 & 1/1 & 1/1 & 1/1 \\
2020AMC10B18 & Probability & 4 & 1 & 5 & 1/1 & 1/1 & 1/1 & 1/1 \\
2021AIMEI12 & Probability & 2 & 3 & 5 & 1/1 & 1/1 & 1/1 & 1/1 \\
2021AMC10A9 & Probability & 1 & 0 & 1 & 1/1 & 1/1 & 1/1 & 1/1 \\
2021AMC12B17 & Probability & 3 & 0 & 3 & 1/1 & 1/1 & 1/1 & 1/1 \\
2023AIMEI6 & Probability & 4 & 0 & 4 & 1/1 & 1/1 & 1/1 & 1/1 \\
2023AMC12A17 & Probability & 4 & 0 & 4 & 1/1 & 1/1 & 1/1 & 1/1 \\
2023AMC12A21 & Probability & 2 & 0 & 2 & 1/1 & 1/1 & 1/1 & 1/1 \\
2024AMC10A17 & Probability & 1 & 1 & 2 & 1/1 & 1/1 & 1/1 & 1/1 \\

%% file: generated_tables/table_A3_rows.tex
overall & all\_80 & Claude & 80 & -1.337 & -1.625 & -1.062 & 0.0 \\
overall & all\_80 & DeepSeek & 80 & -0.812 & -1.038 & -0.588 & 0.0 \\
overall & all\_80 & Gemini & 80 & -0.412 & -0.637 & -0.200 & 0.0008 \\
overall & all\_80 & GPT & 80 & -0.825 & -1.075 & -0.588 & 0.0 \\
domain & Algebra & Claude & 14 & -0.714 & -1.000 & -0.429 & 0.00390625 \\
domain & Algebra & DeepSeek & 14 & -0.857 & -1.214 & -0.500 & 0.00390625 \\
domain & Algebra & Gemini & 14 & -0.429 & -0.786 & -0.071 & 0.109375 \\
domain & Algebra & GPT & 14 & -0.786 & -1.286 & -0.286 & 0.0234375 \\
domain & Combinatorics & Claude & 19 & -1.158 & -1.789 & -0.632 & 0.00067138671875 \\
domain & Combinatorics & DeepSeek & 19 & -0.579 & -1.053 & -0.105 & 0.05078125 \\
domain & Combinatorics & Gemini & 19 & -0.105 & -0.579 & 0.368 & 0.8359375 \\
domain & Combinatorics & GPT & 19 & -0.737 & -1.316 & -0.158 & 0.046630859375 \\
domain & Geometry & Claude & 16 & -2.188 & -2.938 & -1.500 & 6.103515625e-05 \\
domain & Geometry & DeepSeek & 16 & -1.438 & -2.000 & -0.875 & 0.0009765625 \\
domain & Geometry & Gemini & 16 & -0.875 & -1.438 & -0.312 & 0.01904296875 \\
domain & Geometry & GPT & 16 & -1.312 & -1.938 & -0.750 & 0.0009765625 \\
domain & Number Theory & Claude & 15 & -1.733 & -2.333 & -1.200 & 0.000244140625 \\
domain & Number Theory & DeepSeek & 15 & -0.867 & -1.200 & -0.533 & 0.001953125 \\
domain & Number Theory & Gemini & 15 & -0.800 & -1.133 & -0.467 & 0.001953125 \\
domain & Number Theory & GPT & 15 & -1.133 & -1.533 & -0.800 & 0.000244140625 \\
domain & Probability & Claude & 16 & -0.875 & -1.438 & -0.375 & 0.01171875 \\
domain & Probability & DeepSeek & 16 & -0.375 & -0.875 & 0.125 & 0.24365234375 \\
domain & Probability & Gemini & 16 & 0.062 & -0.375 & 0.438 & 1.0 \\
domain & Probability & GPT & 16 & -0.188 & -0.500 & 0.125 & 0.453125 \\